%% file: BuildingNet.tex
\documentclass[10pt,twocolumn,letterpaper]{article}

\usepackage{iccv}
\usepackage{times}
\usepackage{epsfig}
\usepackage{graphicx}
\usepackage{amsmath}
\usepackage{amssymb}

\usepackage{multirow}
\usepackage{comment}
\usepackage{enumitem}
\usepackage{subfigure}
\usepackage{wrapfig}
\usepackage{xcolor}
\usepackage{verbatim}
\usepackage{caption}
\usepackage{etoolbox}

\usepackage[pagebackref=true,breaklinks=true,letterpaper=true,colorlinks,bookmarks=false]{hyperref}

\newcommand\blfootnote[1]{%
  \begingroup
  \renewcommand\thefootnote{}\footnote{#1}%
  \addtocounter{footnote}{-1}%
  \endgroup
}

\iccvfinalcopy

\ificcvfinal\fi

\begin{document}

\input{symbols_format.tex}

\makeatletter
\renewcommand{\paragraph}{%
  \@startsection{paragraph}{4}%
  {\z@}{0.4ex \@plus 1ex \@minus .2ex}{-0.75em}%
  {\normalfont\normalsize\bfseries}%
}
\makeatother

%%%%%%%%% TITLE

\title{BuildingNet: Learning to Label 3D Buildings \vspace{-7mm}}

\author{ 
        Pratheba Selvaraju$^1$
    \and
        Mohamed Nabail$^1$
    \and
        Marios Loizou$^2$
    \and
        Maria Maslioukova$^2$
    \and
        Melinos Averkiou$^2$
    \and
        Andreas Andreou$^2$
    \and
        Siddhartha Chaudhuri$^3$
     \and           
        Evangelos Kalogerakis$^1$
     \and \vspace{-5mm}
     \\ 
    $^1$UMass Amherst                            \,\,\,\,\,\,\,\,
    $^2$University of Cyprus / CYENS CoE Cyprus  \,\,\,\,\,\,\,\,
    $^3$Adobe Research / IIT Bombay
}

\twocolumn[{%
 \renewcommand\twocolumn[1][]{#1}%
 \maketitle
 \thispagestyle{empty}
 \vspace{-8mm}
 \centering
 \includegraphics[width=\textwidth]{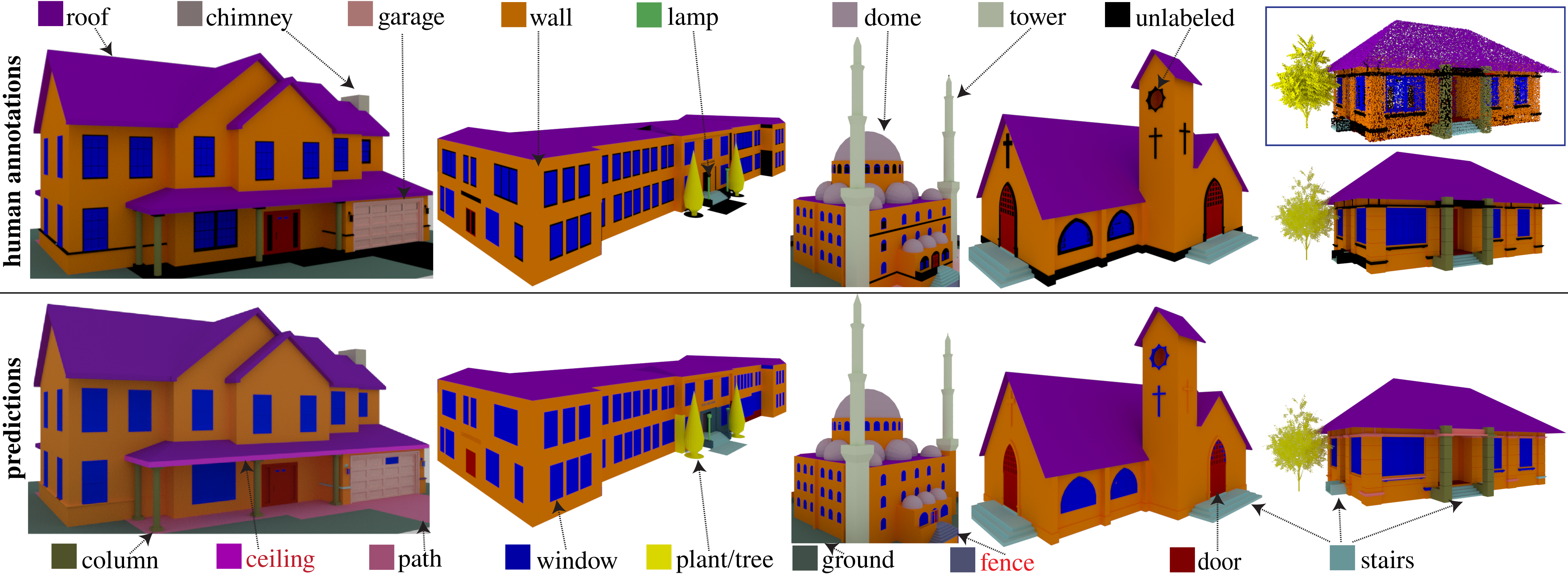}
 \vspace{-7mm}
 \captionof{figure}
  {
  We introduce a dataset of 3D building meshes with annotated exteriors (top). We also present a graph neural network that processes building meshes and labels them by encoding structural and spatial relations between mesh components (bottom). 
  Our dataset also includes a point cloud track (blue box). Examples of erroneous network outputs 
  are in \textcolor[rgb]{0.80,0.20,0.25}  {
  red text.}
  \label{fig:teaser}
   }
\vspace{3mm}
}]

\maketitle

\ificcvfinal\thispagestyle{empty}\fi

%%%%%%%%% ABSTRACT
\begin{abstract}
\vspace{-4mm} We introduce BuildingNet: (a) a large-scale dataset of 3D building models whose exteriors are consistently labeled, and (b) a graph neural network that labels building meshes by analyzing spatial and structural relations of their geometric primitives. To create our dataset, we used crowdsourcing combined with expert guidance, resulting in $513K$ annotated mesh primitives, grouped into $292K$ semantic part components across $2K$ building models. The dataset covers several building categories, such as houses, churches, skyscrapers, town halls, libraries, and castles. We include a benchmark for evaluating mesh and point cloud labeling. Buildings have more challenging structural complexity compared to objects in existing benchmarks (e.g., ShapeNet, PartNet), thus, we hope that our dataset can nurture the development of algorithms that are able to cope with such large-scale geometric data for both vision and graphics tasks e.g., 3D semantic segmentation, part-based generative models, correspondences, texturing, and analysis of point cloud data acquired from real-world buildings. Finally, we show that our mesh-based graph neural network significantly improves performance over several baselines for labeling 3D meshes.
Our project page \url{www.buildingnet.org} includes our dataset and code.

\blfootnote{\hspace{-5mm}\copyright  2021 IEEE. Personal use of this material is permitted. Permission from IEEE must be obtained for all other uses, in any current or future media, including reprinting/republishing this material for advertising or promotional purposes, creating new collective works, for resale or redistribution to servers or lists, or reuse of any copyrighted component of this work in other works. The final published version is available on IEEE Xplore.}
\end{abstract}
\vspace{-7mm}

%%%%%%%%% BODY TEXT

%%%%%%%%% Introduction
\section{Introduction}
\input{introduction.tex}

%%%%%%%%% Related Work
\section{Related Work}
\input{related_work.tex}

%%%%%%%%% Building Data Annotation
\section{Building Data Annotation}
\label{sec:dataset}
\input{dataset.tex}

%%%%%%%%% Building GNN
\section{Building GNN}
\label{sec:net}
\input{net.tex}

%%%%%%%%% Results
\section{Results}
\label{sec:results}
\input{results.tex}

%%%%%%%%% Limitations and Conclusion
\section{Discussion}
\input{conclusion.tex}

%%%%%%%%% References
{\small
\bibliographystyle{ieee_fullname}
\bibliography{BuildNet_bibliography}
}

\input{supplementary.tex}

\end{document}

%% file: symbols_format.tex
%%%%%%%%%%%%%%%%%%%%%%%%%%%%%%%%%%%%%%%%%%%%%%%%%%%%%%%%%%%%%%%%%%%
%%%%%%%%%%%%%%%%%%%%%%% SYMBOLS %%%%%%%%%%%%%%%%%%%%%%%%%%%%%%%
%%%%%%%%%%%%%%%%%%%%%%%%%%%%%%%%%%%%%%%%%%%%%%%%%%%%%%%%%%%%%%%%%%%
\newcommand{\ba}{\mathbf{a}}
\newcommand{\bb}{\mathbf{b}}
\newcommand{\bc}{\mathbf{c}}
\newcommand{\bd}{\mathbf{d}}
\newcommand{\be}{\mathbf{e}}
\newcommand{\bff}{\mathbf{f}}
\newcommand{\bg}{\mathbf{g}}
\newcommand{\bh}{\mathbf{h}}
\newcommand{\bi}{\mathbf{i}}
\newcommand{\bj}{\mathbf{j}}
\newcommand{\bk}{\mathbf{k}}
\newcommand{\bl}{\mathbf{l}}
\newcommand{\bm}{\mathbf{m}}
\newcommand{\bn}{\mathbf{n}}
\newcommand{\bo}{\mathbf{o}}
\newcommand{\bp}{\mathbf{p}}
\newcommand{\bq}{\mathbf{q}}
\newcommand{\br}{\mathbf{r}}
\newcommand{\bs}{\mathbf{s}}
\newcommand{\bt}{\mathbf{t}}
\newcommand{\bu}{\mathbf{u}}
\newcommand{\bv}{\mathbf{v}}
\newcommand{\bw}{\mathbf{w}}
\newcommand{\bx}{\mathbf{x}}
\newcommand{\by}{\mathbf{y}}
\newcommand{\bz}{\mathbf{z}}
\newcommand{\bA}{\mathbf{A}}
\newcommand{\bB}{\mathbf{B}}
\newcommand{\bC}{\mathbf{C}}
\newcommand{\bD}{\mathbf{D}}
\newcommand{\bE}{\mathbf{E}}
\newcommand{\bF}{\mathbf{F}}
\newcommand{\bG}{\mathbf{G}}
\newcommand{\bH}{\mathbf{H}}
\newcommand{\bI}{\mathbf{I}}
\newcommand{\bJ}{\mathbf{J}}
\newcommand{\bK}{\mathbf{K}}
\newcommand{\bL}{\mathbf{L}}
\newcommand{\bM}{\mathbf{M}}
\newcommand{\bN}{\mathbf{N}}
\newcommand{\bO}{\mathbf{O}}
\newcommand{\bP}{\mathbf{P}}
\newcommand{\bQ}{\mathbf{Q}}
\newcommand{\bR}{\mathbf{R}}
\newcommand{\bS}{\mathbf{S}}
\newcommand{\bT}{\mathbf{T}}
\newcommand{\bU}{\mathbf{U}}
\newcommand{\bV}{\mathbf{V}}
\newcommand{\bW}{\mathbf{W}}
\newcommand{\bX}{\mathbf{X}}
\newcommand{\bY}{\mathbf{Y}}
\newcommand{\bZ}{\mathbf{Z}}
\newcommand{\bzero}{\mathbf{0}}
\newcommand{\balpha}{\mbox{\boldmath$\alpha$}}
\newcommand{\bgamma}{\mbox{\boldmath$\gamma$}}
\newcommand{\bGamma}{\mbox{\boldmath$\Gamma$}}
\newcommand{\bmu}{\mbox{\boldmath$\mu$}}
\newcommand{\bphi}{\mbox{\boldmath$\phi$}}
\newcommand{\bPhi}{\mbox{\boldmath$\Phi$}}
\newcommand{\bSigma}{\mbox{\boldmath$\Sigma$}}
\newcommand{\bsigma}{\mbox{\boldmath$\sigma$}}
\newcommand{\btheta}{\mbox{\boldmath$\theta$}}

\newcommand{\mE}{\mathcal{E}}
\newcommand{\mV}{\mathcal{V}}
\newcommand{\mM}{\mathcal{M}}
\newcommand{\mL}{\mathcal{L}}
\newcommand{\mU}{\mathcal{U}}
\newcommand{\mC}{\mathcal{C}}
\newcommand{\mS}{\mathcal{S}}
\newcommand{\mR}{\mathcal{R}}
\newcommand{\mD}{\mathcal{D}}
\newcommand{\mT}{\mathcal{T}}
\newcommand{\mSl}{\mathcal{S}_l}
\newcommand{\mN}{\mathcal{N}}
\newcommand{\mDll}{\mathcal{D}_{l,l'}}

\newcommand{\sdash}{\mbox{-}}

\newcommand{\ra}{\rightarrow}
\newcommand{\la}{\leftarrow}

\def\A{{\cal A}}
\def\B{{\cal B}}
\def\C{{\cal C}}
\def\D{{\cal D}}
\def\E{{\cal E}}
\def\F{{\cal F}}
\def\G{{\cal G}}
\def\H{{\cal H}}
\def\I{{\cal I}}
\def\J{{\cal J}}
\def\K{{\cal K}}
\def\L{{\cal L}}
\def\M{{\cal M}}
\def\N{{\cal N}}
\def\O{{\cal O}}
\def\P{{\cal P}}
\def\Q{{\cal Q}}
\def\R{{\cal R}}
\def\S{{\cal S}}
\def\T{{\cal T}}
\def\U{{\cal U}}
\def\V{{\cal V}}
\def\W{{\cal W}}
\def\X{{\cal X}}
\def\Y{{\cal Y}}
\def\Z{{\cal Z}}
\def\Re{{\mathbb R}}
\def\Cx{{\mathbb C}}
\def\Ze{{\mathbb Z}}
\def\Na{{\mathbb N}}
\def\ud{\mathrm{d}}
\def\eps{\varepsilon}
\def\dist{\textrm{dist}}

%% file: introduction.tex
Architecture is a significant application area of 3D vision. There is a rich body of research on autonomous perception of buildings, led in large part by digital map developers seeking rich annotations and 3D viewing capabilities for building exteriors~\cite{google2017maps}, as well as roboticists who design robots to operate in building interiors (e.g. ~\cite{sepulveda2018nav}). Recent advances in AR/VR also rely on computer-aided building analysis~\cite{Chen:CGF:2020}. Early work on digital techniques for architectural design, including freeform design explorations as well as full-fledged constructions~\cite{arteta2017history}, led to the current ubiquity of computational design tools in architectural studios. In addition, computers can automate the processing of architectural data such as photographs, satellite images and building plans, for archival and analytical purposes (e.g. ~\cite{zeppelzauer2018age,mahmud2020overhead}).

Thus, there is significant incentive to apply modern data-driven geometry processing to the analysis of buildings. However, while buildings are bona fide geometric objects with well-established design principles and clear ontologies, their structural and stylistic complexity is typically greater than, or at least markedly {\em different from}, those of shapes in common 3D datasets like ShapeNet~\cite{Chang:2015} and ScanNet~\cite{Dai:2017}. This makes them challenging for standard shape analysis pipelines, both for discriminative tasks such as classification, segmentation and point correspondences, as well as for generative tasks like synthesis and style transfer. Further, data-driven methods demand data, and to the best of our knowledge there are no large-scale, consistently-annotated, public datasets of 3D building models.

In this paper, we present BuildingNet, the first publicly available large-scale dataset of annotated 3D building models whose exteriors and surroundings are consistently labeled. The dataset provides $513$K annotated mesh primitives across $2$K building models. We include a benchmark for mesh and point cloud labeling, and evaluate 
several mesh and point cloud labeling networks. These methods were developed primarily for smaller single objects or interior scenes and are less successful on architectural data. 

In addition, we introduce a graph neural network (GNN)  that labels building meshes by analyzing spatial and  structural  relations of their geometric primitives. 
 Our GNN treats each subgroup as a node, and takes advantage of relations, such as adjacency and containment, between pairs of nodes. Neural message passing in the graph yields the final mesh labeling. Our experiments show that this approach yields significantly better results for 3D building data than prior methods. 
To summarize, our contributions are:
\begin{itemize}[noitemsep,topsep=0pt,parsep=0pt,partopsep=0pt,leftmargin=22pt]
\item The first large-scale, publicly available 3D building dataset with annotated parts covering several common categories, in addition to a benchmark.
\item A graph neural network that leverages pre-existing noisy subgroups in mesh files to achieve state-of-the-art results in labeling building meshes.
\item An annotation interface and crowdsourcing pipeline for collecting labeled parts of 3D meshes, which could also extend to other categories of 3D data.
\end{itemize}

\begin{figure*}[t!]
   \includegraphics[width=1\textwidth]{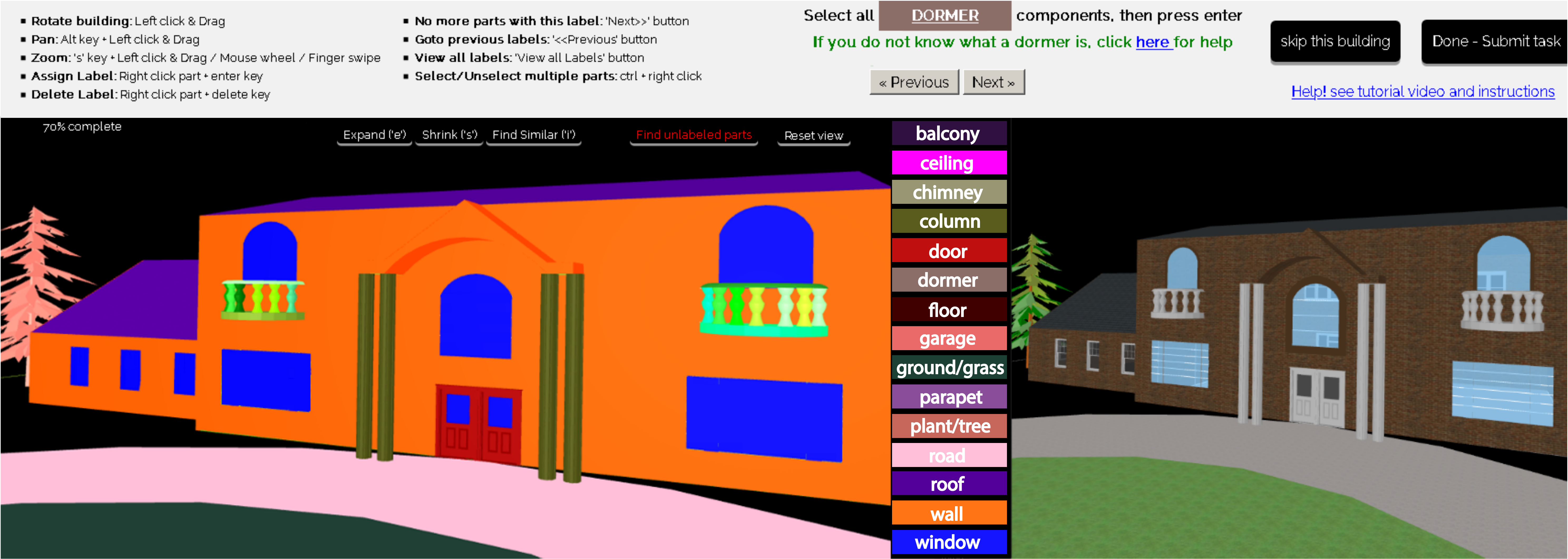}
   \vspace{-6mm}
   \caption{Our interface for labeling 3D building models. The colors of annotated  components  follow the  legend  in the middle (we show  here a subset of labels - the UI\ contained 16 more labels in a more extended layout). Any components that have not been labeled so far are shown in shades of light yellow/green (e.g., balcony components). The UI displays  instructions on top and offers functionality to facilitate labeling, such as automatic detection of repeated components (``find similar''), automatic grouping/un-grouping of components  (``expand''/``shrink''),  and   auto-focusing   on  unlabeled ones (``find unlabeled parts''). }
   \vspace{-3mm}
   \label{fig:interface}
\end{figure*}

%% file: related_work.tex
\paragraph{3D shape semantic segmentation datasets.} Existing  datasets and benchmarks for  3D shape  semantic segmentation are limited to objects with relatively simple structure and small number of  parts~\cite{Chen:2009,Kalogerakis10,Hu:2012,Li:ShapeNet:2016,Mo_2019_CVPR,Yu:2019}. 
The earliest such benchmark~\cite{Chen:2009,Kalogerakis10} had $380$ objects  with few labeled parts per shape.
More recently, Uy et al. \cite{uy-scanobjectnn-iccv19} released a benchmark with 15K scanned objects but focuses on object classification, with part-level segmentations provided only for chairs.
The most recent and largest semantic shape segmentation benchmark of PartNet \cite{Yu:2019} contains $27K$ objects in $24$ categories, such as furniture, tools, and household items. However, even with PartNet's fine-grained segmentation, its categories still have a few tens of labeled parts on average. Our paper introduces a dataset for part labeling of 3D buildings, pushing semantic segmentation to much larger-scale objects with more challenging structure and several tens to hundreds of parts per shape.

\paragraph{3D indoor scene datasets.} Another related line of work has introduced
datasets with object-level annotations in real-world or synthetic 3D indoor environments ~\cite{Hua:2016,Armeni:2016,nguyen2016robust,
Song:2017,Chang:2017,Dai:2017,InteriorNet18,Zheng2019,fu20203dfuture}. In contrast, our dataset focuses on building exteriors, a rather under-investigated domain with its own challenges. While an indoor scene is made of objects, which are often well-separated or have little contact with each other (excluding floors/walls), a building exterior is more like a coherent assembly of parts (windows, doors, roofs) i.e., a single large shape with multiple connected parts, including surroundings (e.g., landscape). Building exteriors share challenges of single-shape segmentation (i.e., segment parts with clean boundaries along   contact areas) as well as scene segmentation (i.e., deal with the large-scale nature of 3D data). Buildings also come in a variety of sizes, part geometry and style \cite{Lun:2015:StyleSimilarity}, making this domain challenging for both shape analysis and synthesis.

\paragraph{3D urban datasets.} With the explosion of autonomous driving applications,  large-scale 3D\ point cloud datasets capturing urban environments have  appeared \cite{Munoz2009,hackel2017isprs,roynard2018parislille3d,behley2019iccv,tan2020toronto3d}. These datasets include  labels such as roads, vehicles, and sidewalks. Buildings are labeled as a single, whole  object. Our dataset contains annotations of building parts, which has its own challenges, as discussed above. The RueMonge14 dataset contains  3D building frontal facades captured from a street in Paris with $8$ labels related to buildings \cite{Riemenschneider14}.
Our buildings are instead complete 3D models with significantly more challenging diversity in geometry, style, function, and with more fine-grained part labels.

\paragraph{Deep nets for 3D mesh understanding.} A few recent neural architectures have been proposed for processing meshes. Some network directly operate on the mesh geometric or topological features ~\cite{masci2015geodesic,Hanocka:2019,Lahav:2020,Schult:2020},
spectral domain \cite{Boscaini2016,Monti2017,Yi2017:syncspec,qiao2019laplaciannet}, while others transfer representations learned by other networks operating, e.g., on mesh views or voxels ~\cite{Kalogerakis:2017:ShapePFCN,Wang:2018,Akundu:2020}.
Our method is complementary to these approaches. It is specifically designed to process meshes with pre-existing structure in the form of mesh components (groups of triangles), which are particularly common in 3D building models. CRFs and various grouping strategies with heuristic criteria have been proposed to aggregate such components into labeled parts \cite{Wang:2018}. Our method instead uses a GNN to label components by encoding spatial and structural relations between them in an end-to-end manner. From this aspect, our method is also related to approaches that place objects in indoor scenes using GNNs operating on bounding box object representations with simple spatial relations,
\cite{Zhou:2019,Kai19}, and GNN approaches for indoor scene parsing based  on graphs defined over point clusters  \cite{Landrieu:2018}. 
Our GNN instead aims to label mesh components represented by rich geometric features, and captures spatial and structural relations specific to building exteriors.

\paragraph{3D Building Mesh Segmentation and Labeling.} 
There has been relatively little work in this area.
Early approaches for semantic segmentation of buildings relied on shallow pipelines with hand-engineered point descriptors and rules \cite{Toshev2010,Martinovic2015}. 
A combinatorial algorithm that groups faces into non-labeled components spanning the mesh with high repetition was proposed in \cite{Demir:2015_2}. A user-assisted segmentation  algorithm was proposed in  \cite{Demir:2015}. Symmetry has   been proposed as  a useful cue to group architectural components \cite{Kobyshev16,Mitra2006}. 
Our method instead aims to label 3D building meshes with a learning-based approach based on 
modern deep backbones for extracting point descriptors. It also incorporates repetitions as a cue for consistent labeling,  along with several other geometric and structural cues.

%% file: dataset.tex
We first discuss the procedure we followed to annotate 3D building models.   
 In contrast to 3D models of small and mid-scale objects, such as tools, furniture, and vehicles encountered in existing 3D\ shape segmentation benchmarks, such as ShapeNet~\cite{Li:ShapeNet:2016,ShapeNetSem} and PartNet~\cite{Mo_2019_CVPR}, buildings tend to contain much richer structure, as indicated by their mesh metadata. For example, one common type of metadata are groupings of polygon faces, commonly known  as \textit{mesh subgroups}~\cite{Mo_2019_CVPR},
which correspond to  geometric primitives and modeling operations used by modelers while designing  shapes. These subgroups often correspond to  ``pieces'' of semantic parts e.g., a window is made of subgroups representing individual horizontal and vertical frame pieces or glass parts. The average  number of mesh subgroups per object at the\ last level of group hierarchy in the largest shape segmentation benchmark (PartNet \cite{Mo_2019_CVPR}) is $24.4$, and the median is $11$. In our dataset, the average number of mesh subgroups per building is $25.5$x larger ($623.6$ subgroups), while the median
is $44$x larger ($497.5$ subgroups). We note that these numbers include only  building exteriors i.e., without considering  building interiors  (e.g, indoor furniture). PartNet relied on  mesh subgroups for faster annotation i.e., the  annotators were manually clicking and grouping them into  parts. Selecting each individual mesh subgroup
in our case would be  too laborious in the case
of a large-scale 3D\ building dataset. To this end, we developed a user interface (UI) that followed the PartNet's principles of  well-defined and consistent labelings, yet its primary focus was to deal with the annotation of a massive number of mesh subgroups per building. In particular, our UI offers annotators the option of \emph{label propagation} to similar subgroups based on both  geometric and mesh metadata
to enable faster labeling. Another  focus was to achieve 
\emph{consensus} across several trained crowdworkers annotating in parallel. To this end, we employed a majority voting process. We   focused on crowdsourcing annotations for
\emph{common part labels} encountered in buildings. In the rest of this  section, we describe our user interface (UI)  for  interactive labeling of 3D buildings  (Section \ref{sec:interface}),
 and the dataset collection process (Section \ref{sec:dataset}).

\subsection{Interface for labeling}
\label{sec:interface}
Our interface is shown in Figure \ref{fig:interface}.
On the left window, we display the building with a distinct color  assigned to each mesh subgroup. When a subgroup is annotated, it changes color
from the default palette (shades of light green and yellow)
to a predetermined, different color according to its label. On the right,  we display the textured version of the building so that crowdworkers also access color cues useful for labeling. The workers have full 3D\ control of viewpoint  (pan, zoom, rotate). Changes on the viewpoint are reflected in both windows. On the top of the interface, we provide instructions and
links with examples of parts from real-world buildings for each label.   The workers are asked to label the mesh subgroups through 
a sequence of questions e.g., ``label all walls'', then ``label all windows'', and so on.  Alternatively, they can skip the questions,  and directly select a desired part label from  the list appearing in the middle of the UI. To perform an assignment of a currently selected label to a mesh subgroup, the workers simply right-click on it and press enter. Alternatively, they can select multiple subgroups and annotate them altogether. All adjacent subgroups with the same label are   automatically merged into a single labeled \textit{component} to decrease the workload of manual merging. We note that we considered the possibility of incorporating mesh cutting tools to split large subgroups into smaller ones for assigning different labels, as done in PartNet \cite{Mo_2019_CVPR}. However, such  tools require reconstruction into watertight meshes, which could not be achieved for most building subgroups due to their non-manifold geometry,  disconnected or overlapping faces, and open mesh boundaries. For the majority of buildings in our dataset, we observed that each subgroup can be assigned with a single part label without requiring further splits. Annotators were also instructed not to label any (rare) subgroups that contained parts with different labels.

\setlength{\columnsep}{8pt}
\begin{wrapfigure}{R}{0.52\linewidth}
 \vspace{-5mm} 
   \includegraphics[width=1\linewidth]{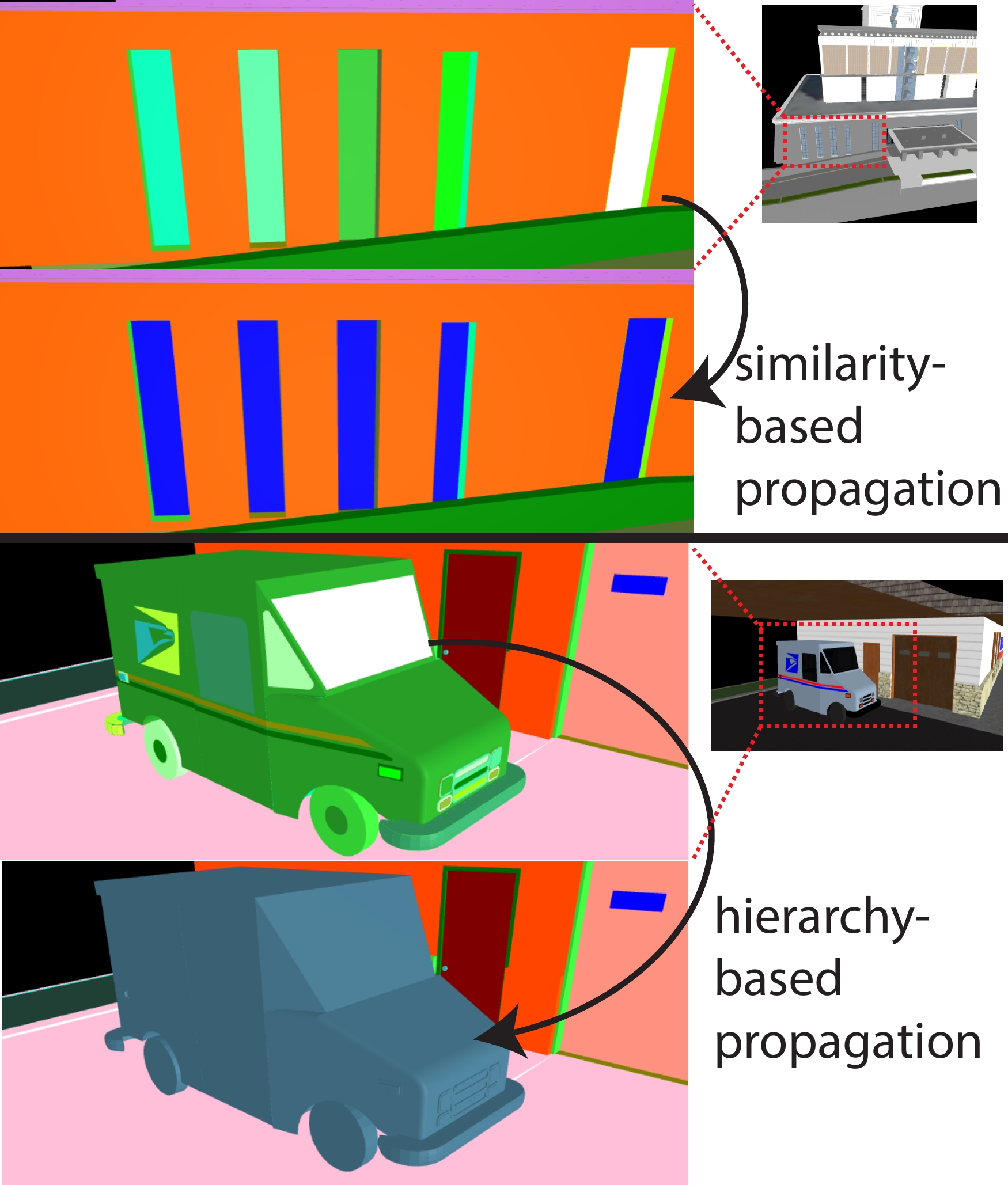}
  \vspace{-7mm}
  \caption{Label propagation to repeated subgroups \emph{(top)} or their parents \emph{(bottom)}. Initially selected subgroup is in white.}
\vspace{-4mm}     
   \label{fig:interface_functionality}
\end{wrapfigure}
Clicking   individual mesh subgroups for assigning part labels can be still cumbersome, since buildings have hundreds or thousands of them. Our UI takes advantage of the fact that buildings often have repeated mesh subgroups e.g., the same window  mesh is re-used multiple times in a facade during 3D\ modeling. Thus, in a pre-processing step, we found all duplicate mesh subgroups by checking if they have  the same mesh connectivity (mesh graph)
and vertex locations match  after factoring out rigid transformations. Details about duplicate detection are provided in the supplementary material (see Appendix A included after the references).
Workers are then given the option to select all subgroup duplicates and propagate the same label to all of them at once, as shown in  Figure \ref{fig:interface_functionality}(top).   
Another UI feature was to allow users to ``expand'' a mesh subgroup selection by taking advantage of any  hierarchical grouping metadata. This expansion was performed by iteratively moving one level up in the mesh group hierarchy and finding all subgroups sharing the same parent with the initially selected subgroup, as shown in  Figure  \ref{fig:interface_functionality}(bottom).
We  refer readers to our supplementary video (see our project website) showing a tutorial with details of our UI operations.

\begin{table}[t!]
\small
\caption{Statistics per building category. \emph{From left to right}: building category, total number of models, average/median/minimum/maximum number of mesh subgroups per model, 
  average number of unique subgroups.}
\vspace{-3mm}
\centering 
\renewcommand{\arraystretch}{0.85}
\begin{tabular}{|@{}c@{}||@{}c@{} |@{}c@{} |@{}c@{} |@{}c@{} |@{}c@{} |@{}c@{} |@{}c@{} |@{}c@{} |@{}c@{} |@{}c@{}| @{}c@{}| @{}c@{}| @{}c@{}| @{}c@{}| @{}c@{}| @{}c@{}| @{}c@{}| @{}c@{}| @{}c@{}| @{}c@{}|}
\hline
\multirow{2}{*}{Category} &
\,num\#\, &
\,avg\#\, & \,med\#\, & \,min\#\, & \,max\#\, &
\,avg\# un.\, \\
& \,models\,
& \,subgrps\, & \,subgrps\, & \,subgrps\, & \,subgrps\, &
  \,subgrps\, \\
\hline 
\hline 
Residential  & 1,424 & 678.7 & 547 & 83 & 1989 & 167.1  \\
\,Commercial\,  & 153 & 723.4 & 606 & 90 & 1981 & 159.8  \\
Religious  & 540 & 487.0 & 348 & 93 & 1981 & 139.9 \\
Civic      & 67 & 628.8 & 480 & 118 & 1822 & 144.4  \\
Castles & 85 & 609.8 & 485 & 125 & 1786 & 193.0 \\
\hline
Whole Set & 2,000 &\ 623.6 & 497.5 & 83 & 1989 & 160.5 \\
\hline 
\end{tabular}
\label{table:basic_stats} 
\vspace{-3mm}
\end{table}

\subsection{Dataset and Benchmark}
\label{sec:dataset}

To create our dataset, we  mined building models from the  3D Warehouse repository~\cite{Web:3DWarehouse}. Mining was driven by various quality checks e.g., excluding low-poly, incomplete, untextured meshes, 
and meshes with no or too few subgroups.
We also categorized them into basic classes following the Wikipedia's article on  ``list of building types''~\cite{Web:Wikipedia} and an Amazon MTurk questionnaire.  
Since we aimed to gather annotations of building exteriors, during a pre-processing step
 we removed interior structure from each building. This was done by performing exhaustive ray casting
originating from  mesh faces of each subgroup and checking if the rays were blocked. We also used ray casting to orient faces such that their normals are  pointing outwards~\cite{Takayama2014Orientation}.
Details about mining, classifying, and pre-processing of the 3D models are given in our supplement.

\paragraph{Part labels.} To determine a set of \emph{common} labels required in our UI to annotate building exteriors, we launched an initial user study involving  a small subset of $100$ buildings across all classes and $10$ participants with domain expertise (graduate students
in   civil engineering and architecture). For this study, we created a variant of our UI asking users to explicitly type tags for mesh subgroups.  We selected a list of $31$ frequently entered tags  to define our label set (see Table \ref{table:label_stats} and Appendix B of our supplement for details).

\paragraph{Annotation procedure.} One possibility to annotate building parts would be to hire ``professionals'' (e.g., architects).  Finding tens or hundreds of such professionals would be   extremely challenging and costly in terms of  time and resources. In an early attempt to do so, we found that consistency was still hard to achieve without additional verification steps and majority voting.
On the other hand,  hiring  non-skilled, non-trained crowdworkers would  have the   disadvantage of gathering erroneous annotations.  We instead proceeded with a more selective approach, where we identified crowdworkers after verifying their ability to conduct the annotation task reliably based on our provided tutorial and instructions. During our worker qualification stage, we released our UI on  MTurk accessible to any worker interested in performing the task. After a  video tutorial, including a web page presenting real-world examples of parts per label, the workers were asked to label a building randomly selected from a predetermined pool of buildings with diverse structure and part labels. 
We then checked their labelings, and qualified those workers whose  labeling was consistent with our instructions.  We  manually verified the quality of their  annotations. Out of $2,520$ participants,  $342$ workers qualified. After this stage, we released our  dataset only to qualified MTurkers. We asked them to label as many parts as they can with a tiered compensation to encourage more labeled area (ranging from $\$0.5$ for labeling  minimum $70\%$ of the building area to $\$1.0$ for labeling $>90\%$ ). Out of the $342$ qualified MTurkers, $168$ accepted to perform the task in this phase.  Each qualified MTurker annotated $\mathord{\sim}60$ buildings and each annotation took $\mathord{\sim}19.5$min on average. 

\setlength{\columnsep}{5pt}
\begin{wrapfigure}{R}{0.5\linewidth}
 \vspace{-4mm} 
   \includegraphics[width=1\linewidth]{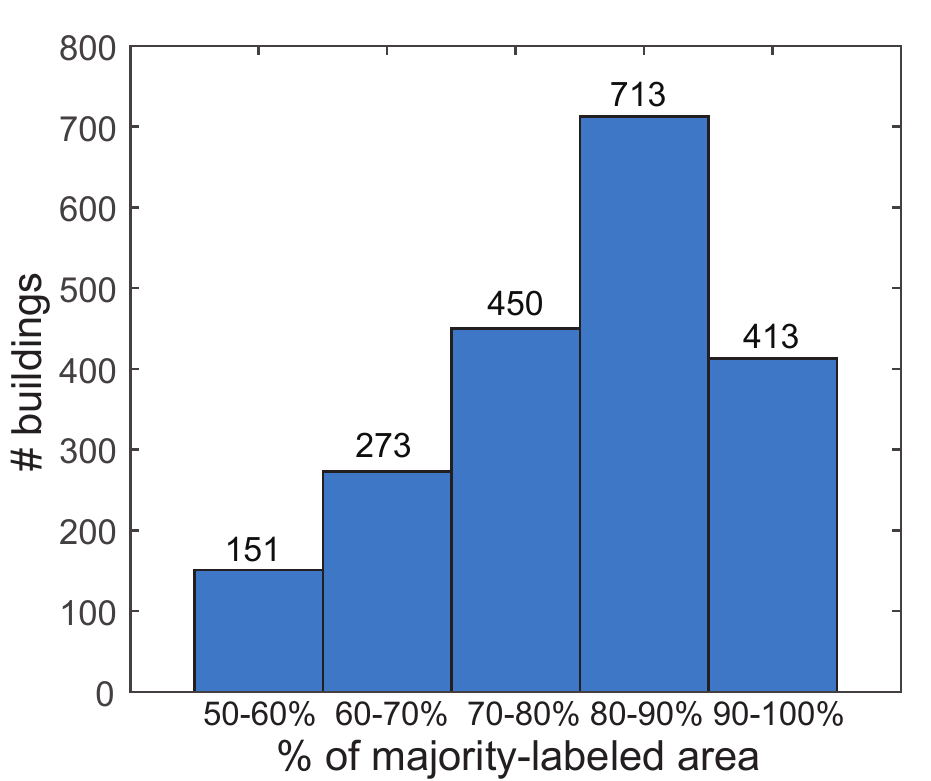}
  \vspace{-7mm}
\vspace{-2mm}     
   \label{fig:distributions_area_comp}
\end{wrapfigure}
\vspace{-1mm}
\paragraph{Dataset.} We gathered
annotations for $2$K buildings. Each building 
was annotated by  $5$ different, qualified MTurkers ($10$K annotations in total). We  accepted a label for each  subgroup if  a majority of  at least $3$ MTurkers out of $5$ agreed on it. 
The inlet figure shows a histogram displaying the distribution of buildings (vertical axis) for different bins of percentage of surface area labeled with achieved majority  (horizontal axis).   
All  buildings in our dataset have labeled area more than $50\%$, and most  have $>80\%$ area labeled. 
In terms of annotator consistency, i.e., the percentage of times that the subgroup label selected by a qualified MTurker agreed with the majority, we found that it is   $92.0\%$,  indicating that the workers were  highly consistent.
Our resulting 2K dataset has $513,087$ annotated mesh subgroups, and $291,998$ annotated components (after merging adjacent subgroups with the same label). The number of unique annotated subgroups and components are $111,832$ and $86,492$ respectively.  Table \ref{table:basic_stats} presents subgroup statistics for each basic building category. Table \ref{table:label_stats} shows labeled component statistics per part label. We include more statistics in the supplement.

\begin{table}[t!]
\small
\caption{Number of labeled components per part label in our dataset, along with their number  and frequency in the training split, hold-out validation, and test split.} 
\vspace{-3mm}
\centering 
\renewcommand{\arraystretch}{0.85}
\begin{tabular}{|@{}c@{}|@{}c@{}|@{}c@{}|@{}c@{}|@{}c@{}|}
\hline 
 \multirow{2}{*}{Label} & \# labeled & \# in training & \# in validation  & \# in test \\
  &  \,comp.\,    &  split (\%)   & split (\%)  & split (\%) \\  [0.5ex]
\hline 
\hline

Window        & 140,972 &    109,218 (47.8\%) &     15,740  (55.1\%) & 16,014  (46.0\%) \\
Plant        & 26,735 &  20,974 (9.2\%) &     1,870  (6.5\%) & 3,891  (11.2\%) \\
Wall        & 22,814 &   18,468 (8.1\%) & 2,270  (7.9\%) & 2,076  (6.0\%) \\
Roof        & 12,881 &   10,342  (4.5\%) &     1,396  (4.9\%) & 1,143  (3.3\%) \\
Banister     & 13,954 &  9,678 (4.2\%) &     1,467  (5.1\%) & 2,809  (8.1\%) \\
Vehicle        & 8,491 & 7,421 (3.2\%) &     716  (2.5\%) & 354  (1.0\%) \\
Door        & 9,417 &    7,363 (3.2\%) &     785  (2.7\%) & 1,269  (3.6\%) \\
Fence        & 5,932 &   5,637 (2.5\%) &     88  (0.3\%) & 207  (0.6\%) \\
Furniture     & 6,282 &  5,000 (2.2\%) &     575  (2.0\%) & 707  (2.0\%) \\
Column        & 6,394 &  4,870 (2.1\%) &     623  (2.2\%) & 901  (2.6\%) \\
Beam        & 6,391 &    4,814  (2.1\%)  &  437 (1.5\%)   &  1,140 (3.3\%) \\
Tower        & 4,478 &   3,873 (1.7\%) &     286  (1.0\%) & 319  (0.9\%) \\
Stairs        & 4,193 &  2,960 (1.3\%) &     472  (1.7\%) & 761  (2.2\%) \\
Shutters        & 2,275 &    1,908 (0.8\%) &     77  (0.3\%) & 290  (0.8\%) \\
Ground        & 2,057 &  1,572 (0.7\%) &     229  (0.8\%) & 256  (0.7\%) \\
Garage        & 1,984 &  1,552 (0.7\%) &     182  (0.6\%) & 250  (0.7\%) \\
Parapet        & 1,986 & 1,457 (0.6\%) &     153  (0.5\%) & 376  (1.1\%) \\
Balcony        & 1,847 & 1,442 (0.6\%) &     199  (0.7\%) & 206  (0.6\%) \\
Floor        & 1,670 &   1,257 (0.5\%) &     205  (0.7\%) & 208  (0.6\%) \\
Buttress        & 1,590 &    1,230 (0.5\%) &     53  (0.2\%) & 307  (0.9\%) \\
Dome        & 1,327 &    1,098 (0.5\%) &     114  (0.4\%) & 115  (0.3\%) \\
Path        & 1,257 &    1,008 (0.4\%) &     113  (0.4\%) & 136  (0.4\%) \\
Ceiling        & 1,193 & 903 (0.4\%) &     111  (0.4\%) & 179  (0.5\%) \\
Chimney        & 1,090 & 800 (0.4\%) &     103  (0.4\%) & 187  (0.5\%) \\
Gate        & 827 & 737 (0.3\%) &     65  (0.2\%) & 25  (0.1\%) \\
Lighting        & 921 & 702 (0.3\%) &     51  (0.2\%) & 168  (0.5\%) \\
Dormer       & 798 &    601 (0.3\%) &     48  (0.2\%) & 149  (0.4\%) \\
Pool        & 742 & 544 (0.2\%) &     78  (0.3\%) & 120  (0.3\%) \\
Road        & 590 & 444 (0.2\%) &     55  (0.2\%) & 91  (0.3\%) \\
Arch        & 524 & 393 (0.2\%) &     11  (0.03\%) & 120  (0.3\%) \\
Awning        & 386 &   295 (0.1\%) &     19  (0.1\%) & 72  (0.2\%) \\
\hline
Total        &  291,998 &     228,561   &     28,591   & 34,846   \\
\hline
\end{tabular}
\label{table:label_stats} 
\vspace{-4mm}
\end{table}

\paragraph{Splits.} 
 We split our dataset into $1600$ buildings for training, $200$ for validation, $200$ for testing ($80/10/10\%$ proportion). The dataset has no duplicate buildings. We created the splits such that (a) the distribution of building classes and parts is similar across the splits (Table \ref{table:label_stats} and supplementary) and (b) test buildings  have high majority-labeled area ($>85\%$) i.e., more complete labelings for evaluation.

\paragraph{Tracks.} We provide two tracks in our benchmark. In the first track, called ``BuildingNet-Mesh'', algorithms can access the mesh data, including subgroups. In this aspect, they can take advantage of any pre-existing mesh structure common in 3D building models. The algorithms are evaluated in two conditions: when the RGB texture is available, and when it is not. In the second condition, algorithms must label the building using only geometric information. The second track, called ``BuildingNet-Points'', is designed for large-scale point-based processing algorithms that must deal with  unstructured point cloud data without access to mesh structure or subgroups, which is still challenging even in the noiseless setting.
To this end, for each mesh, we sample $100K$ points with Poisson disc sampling, to achieve a near-uniform sampling similarly to 
PartNet \cite{Mo_2019_CVPR}.
The point normals originate from triangles. There are  also two evaluation conditions: with and without RGB color for points.

%% file: net.tex
We now describe a graph neural network for labeling 3D meshes by taking advantage of pre-existing mesh structure in the form of subgroups. The main idea of the network is to take into account  spatial and structural relations between subgroups to promote more coherent mesh labeling. The input to our network is a 3D building mesh with subgroups $\mC=\{c_i\}_{i=1}^N$, where $N$ is the number of subgroups, and the output is a label per subgroup. 
In the next section, we  describe how the graph representing a building  is created, then we discuss our GNN architecture operating on this graph.

\paragraph{Graph Nodes.} For each 3D building model, we create a node for each mesh subgroup. Nodes carry an initial raw  representation of the subgroup. Specifically, we first sample the mesh with 100K points (same point set used in  the
 ``BuildingNet-Points'' track), then process them through the 3D\ sparse convolutional architecture of Minkowski network (MinkowskiUNet34 variant \cite{choy20194d}). We also experimented using PointNet++ \cite{qi2017pointnetplusplus}. We extract per-point features  from the last layer of these nets, then perform average pooling over the  points originating from the faces of the subgroup to extract an initial node  representation. We   concatenate this representation with the 3D barycenter position  of the subgroup, its mesh surface area, and the coordinates of the  opposite  corners of its   Oriented Bounding Box (OBB) so that we  capture its spatial dimensions explicitly.  The combination of the above features in the resulting $41$D node representation $\bn_i$ yielded better performance in our experiments.    

\paragraph{Proximity edges.} Driven by the observation that nearby subgroups tend to have the same label (e.g., adjacent pieces of glass or frame are labeled as ``window''), or related labels (e.g., windows are often adjacent to walls), we create edges for pairs of subgroups that capture their degree of proximity. To avoid creating an overly dense graph, which would pose
excessive
memory overheads for the GNN,
we created edges for pairs of subgroups whose distance was up to $10\%$ of the average of their  OBB diagonals.
Relaxing this bound did not improve results. To avoid
a hard
dependency on a single threshold, and to capture the degree of subgroup proximity at multiple scales, we computed the percentage of  point samples of each subgroup whose  distance to the other subgroup is 
less than $1\%$, $2.5\%$, $5\%$, and $10\%$ of the average of their OBB diagonals. Given a pair of  subgroups $(c_i,c_j)$, this results in a $4D$ edge raw representation $\be_{i,j}^{(prox)}$, where each entry approximates  the surface area
 percentage of  $c_i$  proximal to $c_j$ at a different scale. 
 Similarly, we compute a $4D$ representation $\be_{j,i}^{(prox)}$ for the opposite edge direction.

\paragraph{Support edges.} Certain arrangements of labels are often expected along the upright axis of the building e.g., the roof is on top of walls. We create a ``supporting''  edge for each subgroup found to support another subgroup, and ``supported-by'' edges of opposite direction for each subgroup found to be supported by another subgroup. The edges are created by examining OBB\ spatial relations. Specifically, as in the case of proximity edges, we compute a multi-scale 4D edge raw representation $\be_{i,j}^{(ontop)}$  measuring the area percentage of $c_i$'s bottom OBB face lying above the  $c_j$'s  top OBB face for different distances  $1\%$, $2.5\%$, $5\%$, $10\%$ of the average of the two OBB's heights. We also compute  a 4D\ edge raw representation  $\be_{j,i}^{(below)}$ corresponding to the the surface area percentage of $c_j$'s  top OBB face 
lying beneath the  $c_i$'s  bottom OBB face.

\paragraph{Similarity edges.}  Subgroups placed under a symmetric arrangement often share the same label (e.g., repeated windows along a facade). We create an edge per pair of subgroups capturing repetition. For each pair of subgroups, we compute the bidirectional Chamfer distance between their sample points after rigid alignment. To promote robustness to any minor misalignment, or small geometric differences between subgroups, we create similarity edges if the Chamfer distance $d_{i,j}$ is less than $10\%$ of the average of their OBB  diagonals. Increasing this bound did not improve results. We normalize it within $[0,1],$  where $1.0$ corresponds to the above upper bound, and use
 $\be_{i,j}^{(symm)}=1-d_{i,j}$ as raw similarity edge representation.
 We also use the same representation for this opposite direction: \mbox{$\be_{j,i}^{(symm)}=\be_{i,j}^{(symm)}$}.

\paragraph{Containment edges.} Driven by the observation that parts, such as doors or windows, are enclosed by, or contained within other larger parts, such as walls, we create edges for pairs of subgroups capturing their degree of containment.
 For each pair of subgroups, we measure the amount of $c_i$'s volume  contained within the $c_j$'s OBB and also their volume Intersection over Union 
 as a 2D edge representation $\be_{i,j}^{(contain)}$ (and similarly for the opposite edge direction).   

\begin{figure}[t!]
\centering
   \includegraphics[width=0.93
   \columnwidth]{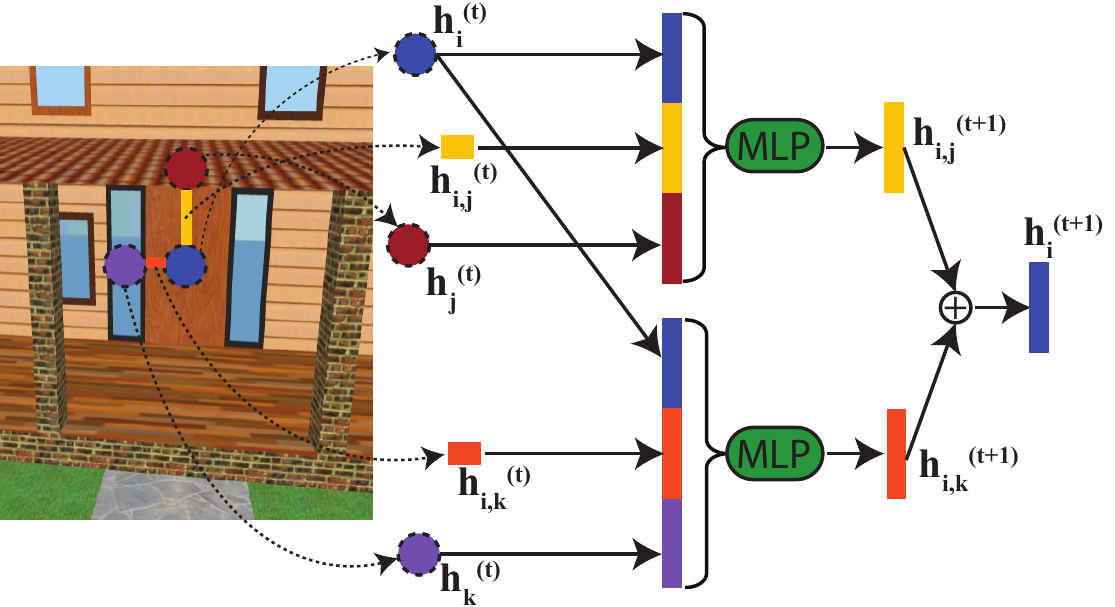}
   \vspace{-3mm}
   \caption{Architecture of the message passing layer. The 
   \textcolor[rgb]{.22,.33,.64}
   {door representation (blue node)} is updated from a
   \textcolor[rgb]{1.0,0.7,0}  {
   support edge (yellow edge)}
   to a 
   \textcolor[rgb]{.62,.13,.15}  {
   roof component (red node)}   
   and a
   \textcolor[rgb]{.95,.35,.16}{
   proximity edge (orange edge)}
   to a 
   \textcolor[rgb]{.47,.36,.65}
   {window (purple node)}.
   }
   \vspace{-4mm}
   \label{fig:GNN_architecture}
\end{figure}

\paragraph{Network architecture.}
 The network updates node and edge representations at each layer inspired by neural message passing~\cite{kipf2018neural}.
 Figure~\ref{fig:GNN_architecture} shows one such layer of message passing.
 Below we explain our architecture at test time. 
 
\paragraph{Initialization.} Given a pair of subgroups $c_i$ and $c_j$, we first concatenate their  edge representations across all types: 
\vspace{-1mm}
\begin{equation*}
\be_{i,j} = \{\be_{i,j}^{(prox)}, \be_{i,j}^{(ontop)}, \be_{i,j}^{(below)},\be_{i,j}^{(contain)},
\be_{i,j}^{(sim)}  \}
\end{equation*}
We note that some  of the edge types might not be present between  two subgroups based on our graph construction.  The entries of our edge representations indicate  degree of proximity, support, containment, or similarity, and are normalized between $[0,1]$ by definition. 
Zero values for an edge representation of a particular type indicate non-existence for this type.
Each  raw edge representation $\be_{i,j}$ is initially processed by a MLP to output a learned  representation $\bh^{(0)}_{i,j}=MLP \big( \be_{i,j}; \bw^{(0)}  \big)$, where $\bw^{(0)}$  are learned MLP\ parameters. The initial node representation is $\bh^{(0)}_{i}=\bn_i$.

\paragraph{Node and edge updates.} Each of the following layers process the node and edge representations of the previous layer  through MLPs and mean aggregation respectively:
\begin{align*}
&\bh^{(l+1)}_{i,j}=MLP \big( \bh^{(l)}_i, \bh^{(l)}_j, \bh^{(l)}_{i,j}; \bw^{(l)}  \big)  \\
& \bh^{(l+1)}_{i}=\frac{1}{|N(i)|} \sum_{j \in \mN(i)} \bh^{(l+1)}_{i,j}
\end{align*}
where $\bw^{(l)}$ are learned MLP\ parameters. We use $3$  layers of node/edge updates. Finally, the last GNN layer processes the  node representations
of the third layer, and decodes them to a probability per label using a MLP and softmax. Details about the architecture are in the supplement.

\paragraph{Training loss.} Since some parts are more rare than others, as shown in Table \ref{table:label_stats}, we use a weighted softmax loss to train our network, where weights are higher for rarer parts to promote correct labeling for them (i.e., higher mean Part IoU). For each building, the loss is $L=-\sum_{c_i \in \mL}  w_l \cdot \hat{\bq}_i \log \bq_{i}$, where $\mL$ is the set of all annotated subgroups in the building, $\hat{\bq}_i$ is the ground-truth one-hot label vector for subgroup $c_i$,  $\bq_i$ is its predicted label probabilities, and $w_l$ is the weight for the label empirically set to be the log of inverse label frequency
(i.e., a smoothed version of inverse frequency weights similarly to \cite{Mikolov}).
 We use the same loss to train the MinkowskiNet used in our node representation: the loss is simply applied to points instead of subgroups. We experimented with other losses, such as the focal loss \cite{focal} and the class-balanced loss \cite{cui2019}, but we did not find  significant improvements in our dataset (see supplementary material).

\paragraph{Implementation details.} Training of the BuildingGNN is done through the Adam optimizer \cite{Kingma15} with learning rate $0.0001$, beta coefficients are $(0.9, 0.999)$ and weight decay is set to $10^{-5}$. We pick the best model and hyper-parameters based on the performance in the holdout validation split.

%% file: results.tex
\begin{figure*}[t!]
   \includegraphics[width=1\textwidth]{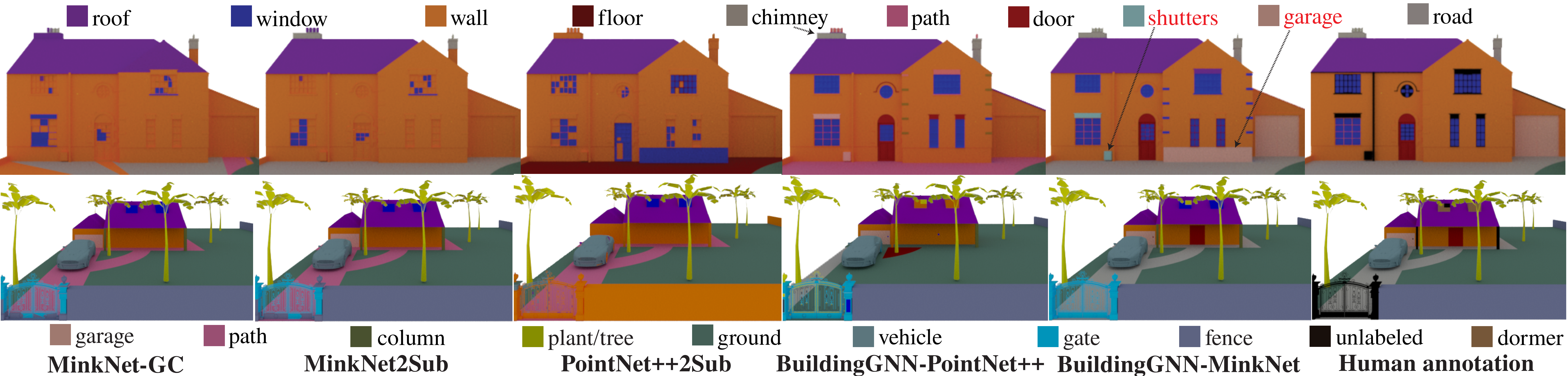}
   \vspace{-6mm}
   \caption{Comparisons with other methods. Despite a few errors (red text), the BuildingGNN is closer to human annotations.}
   \vspace{-3mm}
   \label{fig:comparisons}
\end{figure*}

We now discuss our evaluation protocol, then show qualitative and quantitative   results for our benchmark tracks.

\paragraph{Evaluation protocol.}  Since
most part classes are commonly encountered across different building categories (e.g.,
walls, doors, windows), all evaluated methods are trained across all five building categories (i.e., no category-specific training). Methods must also deal with the part class imbalance of our dataset. For
evaluation in the point cloud track (``BuildingNet-Points''), we use the metrics of  mean shape IoU\ and part IoU, as in PartNet \cite{Mo_2019_CVPR}. We also report the per-point classification accuracy. For the mesh track
(``BuildingNet-Mesh''), the same  measures are applied on triangles. However, since  triangles may  differ in area, we propose the following IoU variations, where the contribution of each triangle is weighted by its face area. Given   all the annotated triangles  across all buildings of the test dataset $T_D$, the part IoU for a label $l$ is measured as:
\begin{equation*}
IoU(l) =  \frac{ \sum_{t \in T_D} a_t \cdot ([y_t==l] \wedge [\hat y_t==l])}
               { \sum_{t \in T_D} a_t \cdot ([y_t==l]   \lor [\hat y_t==l])}
\end{equation*}
where $\hat y_t$ is the majority-annotated (ground-truth) label for a triangle \mbox{$t \in T_d$}, $y_t$ is the predicted label for it, and $[\cdot]$ evaluates the above binary expressions. The shape IoU for a shape $s$ with a set of annotated triangles $T_s$  is measured as:
\begin{equation*}
IoU(s) =  \frac{1}{|L_s|}
\sum_{l \in L_s}  
\frac{ \sum_{t \in T_s} a_t \cdot ([y_t==l] \wedge [\hat y_t==l])}
     { \sum_{t \in T_s} a_t \cdot ([y_t==l]   \lor [\hat y_t==l])}
\end{equation*}
where $L_s$ is the set of all labels present  in the annotations or predictions for that shape.
We also report
the per-triangle classification accuracy weighted by face  area \cite{Kalogerakis10}.
 
\paragraph{``BuildingNet-Points'' track.} As an initial seed for the leaderboard of this track, we evaluated three popular nets  able to handle our $100K$ point sets: PointNet++ \cite{qi2017pointnetplusplus}, MID-FC \cite{Wang-2020-Unsupervised}, and MinkowskiUNet34  \cite{choy20194d}. 
We also tried other point-based networks e.g., DGCNN \cite{dgcnn}, but were unable to handle large point clouds due to excessive memory requirements (see our supplementary material for more discussion). All networks were trained under the same augmentation scheme (12 global rotations per building and small random translations). For all networks, we experimented with SGD, Adam \cite{Kingma15}, with and without warm restarts \cite{loshchilov2017sgdr}, and selected the best scheduler and hyperparameters for each of them based on the validation split. We did not use any  form of pre-training. Table \ref{table:point_cloud_track} reports the results. 
We observe that the MinkowskiNet offers the best performance. We also observe that the inclusion of color  tends to improve performance e.g., we observe a $3\%$ increase in Part IoU for MinkowskiNet. Another observation is that compared to PartNet classes, where the  Part IoU 
ranges between $\mathord{\sim}30-70\%$  for PointNet++, the performance in our dataset is much lower:  PointNet++ has $14.1\%$ Part IoU. Even for the best performing method (MinkowskiNet), the part IoU is still relatively low ($29.9\%$), indicating that our building dataset is substantially more challenging.
\begin{table}[t!]
\small
\caption{``BuildingNet-Point'' track results. The column `$\bn$?' means whether networks use point normals, and the column `$\bc$?' means whether they use RGB color as input.} 
\vspace{-3mm}
\centering 
\renewcommand{\arraystretch}{0.85}
\begin{tabular}{|c|@{}c@{}|@{}c@{}|c|c|c|c|}
\hline
Method  &  \,$\bn$?\, & \,$\bc$?\, & Part IoU  & Shape IoU & Class acc. \\
\hline
\hline
PointNet++       &  $\checkmark$  & $\times$  & 8.8\%  & 12.2\% & 52.7\%\\
MID-FC(nopre)    & $\checkmark$  &   $\times$  &20.9\%  & 19.0\% & 59.4\%\\
MinkNet     & $\checkmark$  &  $\times$ & \textbf{26.9\%} & \textbf{22.2\%} & \textbf{62.2\%} \\
\hline
\hline
PointNet++      & $\checkmark$  & $\checkmark$  & 14.1\%   & 16.7\% & 59.5\%\\
MID-FC(nopre)      & $\checkmark$  & $\checkmark$  & 25.0\%   & 22.3\% & 63.2\%\\
MinkNet       & $\checkmark$  & $\checkmark$    & \textbf{29.9\%}  & \textbf{24.3\%} & \textbf{65.5\%} \\

\hline 
\end{tabular}
\label{table:point_cloud_track} 
\vspace{-3mm}
\end{table}

\paragraph{``BuildingNet-Mesh'' track.} For our mesh track, we first include a number of baselines which rely on networks trained on the point cloud track, then transferring their results to  meshes. One strategy for this transfer is to build correspondences between mesh faces and nearest points. Specifically, for each point we find its nearest triangle. Since some triangles might not be associated with any points, we also build the reverse mapping: for each triangle, we find its closest point. In this manner, every triangle $t$ has a set of points $P_t$ assigned to it with the above bi-directional mapping. Then we perform average pooling of the point probabilities per triangle: $\bq_t = \sum_{p \in P_t} \bq_p / |P_t| $ where $\bq_p$ and $\bq_t$ are point and triangle probabilities respectively. We report results of these baselines
in Table \ref{table:mesh_track}. We note that we tried max pooling, yet average pooling had better performance (see supplement). 
Another strategy is to aggregate predictions based on mesh subgroups instead of triangles i.e., average probabilities of points belonging to each subgroup. This strategy takes advantage of mesh structure and improves results. Another baseline is Graph Cuts (GC) on the mesh, which has been used in mesh segmentation \cite{Kalogerakis10} (see supplement for the GC energy). 
Finally, we report results from our GNN (``BuildingGNN''), using  PointNet++ or MinkowskiNet node features. The BuildingGNN significantly improves the respective baselines e.g., with color as input, BuildingGNN with PointNet++ features improves Part IoU by $15.4\%$ over the best PointNet++ variant, while BuildingGNN with MinkowskiNet features improves Part IoU by $5.6\%$ over the best MinkowskiNet variant. The BuildingGNN with MinkowskiNet features performs the best  with or without color. Our supplement includes an ablation study showing that each edge type in the BuildingGNN improves performance over using node features alone, while the best model is the one with all edges.

\begin{table}[t!]
\small
\caption{``BuildingNet-Mesh'' results. PointNet++2Triangle means triangle-pooling with PointNet++ (similarly for others). PointNet2Sub means subgroup-pooling. MinkNet-GC means graph cuts with MinkowskiUNet34 unary terms.}
\vspace{-3mm}
\centering 
\renewcommand{\arraystretch}{0.85}
\begin{tabular}{|@{}c@{}|@{}c@{}|@{}c@{}|@{}c@{}|@{}c@{}|@{}c@{}|@{}c@{}|@{}c@{}|}
\hline
Method  &  \,$\bn$?\, & \,$\bc$?\, & \,Part IoU\,  & Shape IoU & Class acc. \\
\hline 
\hline 
\,PointNet++2Triangle\,  &  $\checkmark$  & $\times$  & 8.8\%  & 13.1\% & 54.7\% \\
MidFC2Triangle   &  $\checkmark$  & $\times$  & 23.1\%  & 22.1\% & 42.9\%\\
MinkNet2Triangle   &  $\checkmark$  & $\times$  & 28.8\%  & 26.7\% & 64.8\% \\
PointNet++2Sub      &  $\checkmark$  & $\times$  & 9.5\%  & 16.0\% & 57.9\% \\
MidFC2Sub       &  $\checkmark$  & $\times$ & 26.4\%  & 28.4\% & 46.2\%\\
MinkNet2Sub       &  $\checkmark$  & $\times$ &  33.1\%  & 36.0\% & 69.9\% \\
MinkNet-GC        &  $\checkmark$  & $\times$ &  29.9\%  & 28.3\% & 66.0\% \\
BuildingGNN-PointNet++ &  $\checkmark$  & $\times$ &  29.0\%  & 33.5\% & 67.9\%\\
BuildingGNN-MinkNet  &  $\checkmark$  & $\times$ & \textbf{40.0\%}  & \textbf{44.0\%} & \textbf{74.5\%} \\
\hline
\hline
PointNet2Triangle  &  $\checkmark$  & $\checkmark$ &  14.0\%  & 18.0\% & 60.7\%\\
MidFC2Triangle   &  $\checkmark$  & $\checkmark$ &  27.3\%  & 26.2\% & 45.6\%\\
MinkNet2Triangle   &  $\checkmark$  & $\checkmark$ &  32.8\%  & 29.2\% & 68.1\% \\
PointNet2Sub     &  $\checkmark$  & $\checkmark$ &  16.1\%  & 23.5\% & 64.8\%\\
MidFC2Sub      &  $\checkmark$  & $\checkmark$ &  30.3\%  & 33.1\% & 48.6\%\\
MinkNet2Sub      &  $\checkmark$  & $\checkmark$ &  37.0\%  & 39.1\% & 73.2\% \\
MinkNet-GC        &  $\checkmark$  & $\checkmark$ &  33.8\%  & 31.1\% & 68.9\% \\
BuildingGNN-PointNet++ & $\checkmark$  & $\checkmark$ &  31.5\%  & 35.9\% & 73.9\%\\
BuildingGNN-MinkNet &  $\checkmark$  & $\checkmark$ &  \textbf{42.6\%}  & \textbf{46.8\%} &\textbf{77.8\%} \\
\hline
\end{tabular}
\label{table:mesh_track} 
\vspace{-3mm}
\end{table}

\paragraph{Qualitative results.} Figure \ref{fig:comparisons} shows comparisons of BuildingGNN with other methods. We observe that its predictions are closer to human annotations compared to others.
Figure \ref{fig:teaser} presents more results from BuildingGNN.

%% file: conclusion.tex
\vspace{-1mm} We presented the first large-scale dataset for labeling 3D buildings and a GNN that takes advantage of mesh structure to improve labeling. A future avenue of research is to automatically discover segments in point clouds and embed them into a GNN like ours. Currently, edges are extracted heuristically. Learning edges and features in an end-to-end  manner may improve results. Finally, mesh cutting and hierarchical labeling can lead to richer future dataset versions. 

\paragraph{Acknowledgements.} We thank Rajendra Adiga, George Artopoulos, Anastasia Mattheou, Demetris Nicolaou for their help. Our work was funded by Adobe, NSF (CHS-1617333), the ERDF and the Republic of Cyprus through the RIF (Project EXCELLENCE/1216/0352), and the EU H2020 Research and Innovation Programme and the Republic of Cyprus through the Deputy Ministry of Research, Innovation and Digital Policy (Grant Agreement 739578).

%% file: supplementary.tex
\begin{figure}[t!]
   \includegraphics[width=1\columnwidth]{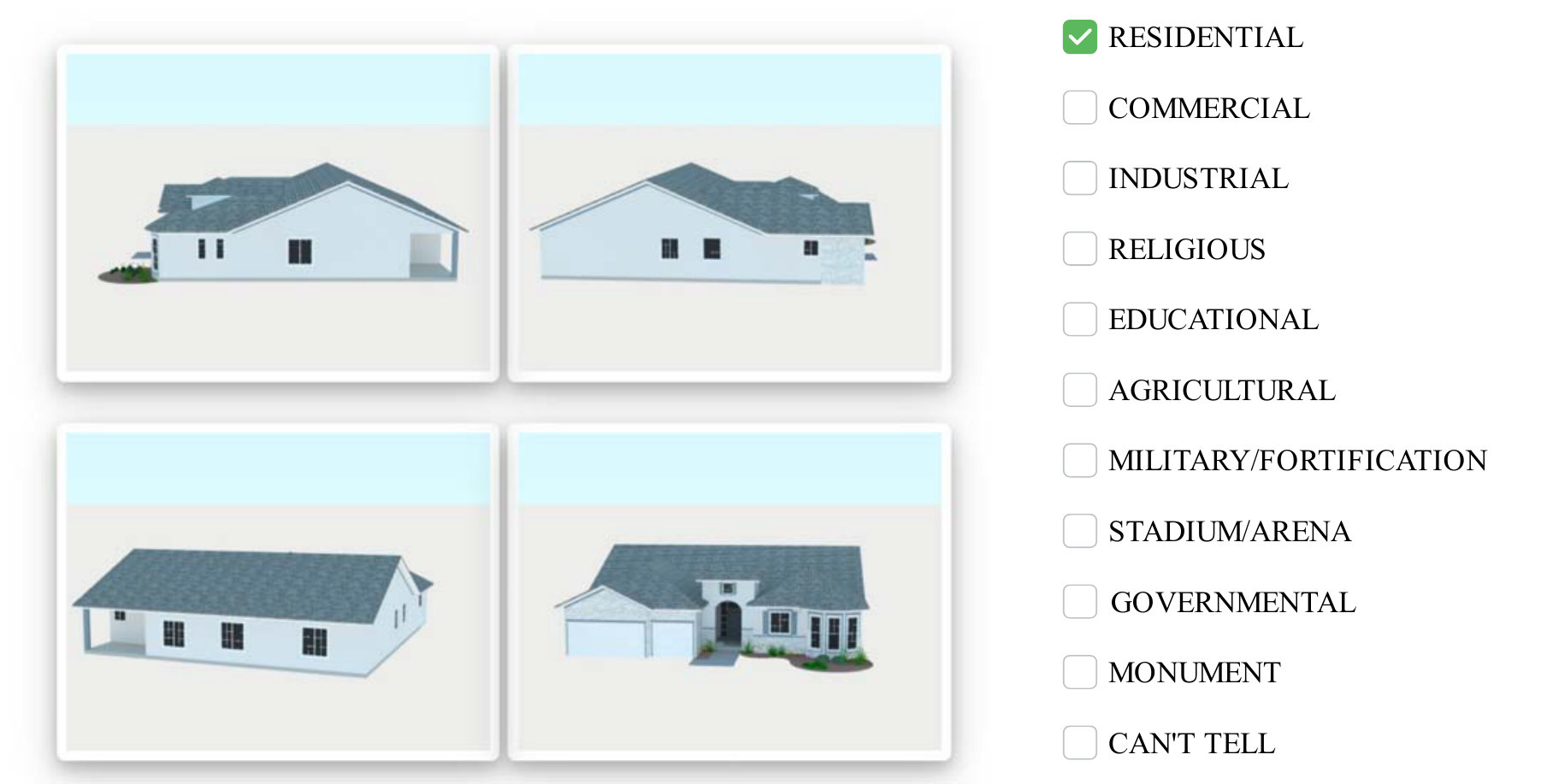}
   \caption{Web questionnaire for classifying a model into basic building categories}
   \label{fig:building_classification_study}
\end{figure}

\newpage
\begin{center}
\section*{-- Supplementary Material --}
\end{center}

\section*{Appendix A: Building collection }
\vspace{2mm}

\paragraph{Mining building models.} We used the Trimble 3D Warehouse repository \cite{Web:3DWarehouse} to mine 3D building models. Specifically, we used keywords denoting various building categories,
following a snapshot from Wikipedia's article on  ``list of building types'' \cite{Web:Wikipedia}. The article contained $181$ common building types, such as  ``house'',  ``hotel'', ``skyscraper'',  ``church'', ``mosque'', ``city hall'', ``castle'',  ``office building'', and so on, organized into basic categories, such as residential, commercial,  industrial, agricultural, military, religious, educational, and governmental
buildings. For each keyword, we retrieved the first $10K$ models. Since some keyword searches returned much fewer buildings, and  since identical models were retrieved across different searches (e.g., a building can have both tags  ``house'' and  ``villa''), we ended up with  $48,439$ models. The models were stored in the COLLADA file format.

\paragraph{Mesh-based filtering.} Low-poly meshes often represent low-quality or incomplete buildings, and they often cause problems in rendering and geometry processing. Thus, we  removed models with less than $3K$ faces and also removed models with extremely large number of faces (more than $1M$ faces) that tend to significantly slow down
mesh processing and  rendering for interactive segmentation (total $13,628$ models removed). Since our UI relies on  labeling mesh subgroups (submeshes) stored in the leaf nodes of the COLLADA hierarchy, we  excluded  under-segmented models with less than $50$ mesh subgroups, and over-segmented models with more than $5K$ mesh subgroups, which would be more challenging to label  (total $4,958$ models removed). 
As a result,  the filtered dataset contained $29,853$ models.

\paragraph{Crowdworker-based filtering.}  The above keyword searches can be affected by noisy metadata, such as erroneous and irrelevant tags not describing  the actual shape class. As a result, most of  the retrieved models did not represent buildings. Some models
 also contained entire neighborhoods or multiple buildings. Thus, our next step was to filter 3D models that did not represent single buildings. We resorted to crowdworkers from
Amazon Mechanical Turk (MTurk) to verify whether each model is a single building or not, and also classify it into basic categories following Wikipedia's  categorization. To this end, we created web questionnaires showing each model from four viewpoints with elevation $0$ degrees from the ground plane, and azimuth difference $90^o$ degrees. We asked MTurk participants (MTurkers) to select a category that best describe the model (see Figure \ref{fig:building_classification_study} for an example of rendered views, and basic categories we used). We instructed them to answer  ``can't tell'' if the displayed model did not represent a single building, or  when they could not recognize it.  

Each participant was asked to complete a questionnaire with $20$ queries  randomly picked from our filtered set of models. Each query showed one model (Figure \ref{fig:building_classification_study}). Queries were shown in a random order. Each query was repeated twice in the questionnaire in a random order to detect unreliable participants providing inconsistent answers (i.e., we had $10$ unique queries per questionnaire).  We filtered out unreliable MTurk participants who gave two inconsistent answers to more than $3$ out of the $10$ unique queries in the questionnaire. Each participant was allowed to answer one questionnaire at most  to ensure participant diversity. We had total $4,344$ different, reliable MTurk participants in this study. For each of the models, we gathered consistent votes from $7$ different MTurk participants. We accepted a building category for a model, if it was voted by at least $5$ out of $7$ MTurkers.
We note that this majority is statistically significant: given $10$ categories, the probability of a model getting $5$ out of $7$ votes given  random answers is negligible according to a binomial test ($p<0.001$).
 We removed models lacking majority votes (i.e., they were not buildings, or the category could not be determined with high agreement).

The categories ``agricultural'', ``industrial'', ``stadium'' 
had  less than $40$ buildings, thus, we decided to exclude them since their part variability and corresponding  labels, would not be sufficiently represented in training, validation, and test splits of the  segmentation dataset. We also decided to merge the ``educational'' and ``governmental'' buildings into a single broader category, called ``civic'' buildings commonly used to characterize both  types of buildings, 
since we observed that the exterior of a governmental building (e.g., town hall) is often similar to the exterior of an educational one (e.g., public library or college).
The remaining number of models characterized as buildings from our study was $2,286$. We note that all models in our dataset are stored as COLLADA files, and have hierarchy tree depth 
$\ge2$ (excl. the root). We refer the reader to Table \ref{table:filted_dataset} for statistics per basic category in our dataset and its splits.

\begin{table}[t!]
\small
\caption{From left to right:\ number of models per basic building category after filtering (original buildings),  number of buildings whose parts were  labeled by crowdworkers in our dataset (labeled buildings), number and percentage of training, hold-out validation and test buildings} 

\vspace{-4mm}
\centering 
\begin{tabular}{|@{}c@{}|@{}c@{}|@{}c@{}|@{}c@{}|@{}c@{}|@{}c@{}|}
\hline
\multirow{2}{*}{Category} & \,\#orig.\, & \,\#label.\, &\ \,\# train.\, & \# \,val. & \,\# test\,\\
  &  \,build.\, &  \,build.\,   &  (\%)   & (\%)  &  (\%) \\  [0.5ex]
\hline 
\hline 
Residential        & 1424 & 1266 & 1007 (62.9\%)  & 133 (66.5\%) & 126 (63.0)\% \\
Commercial         & 153 & 131 & 104 (6.5\%)  & 16 (8.0\%) & 11 (5.5\%) \\
Religious          & 540 & 469 & 386 (24.1\%)  & 38 (19.0\%) & 45 (22.5\%) \\
Civic             & 67 & 61 & 45 (2.8\%)  & 8 (4.0\%) & 8 (4.0\%) \\
Castles            & 85 & 73 & 58 (3.6\%)  & 5 (2.5\%) & 10 (5.0\%) \\
\hline
Total: & 2286 & 2,000  & 1600 (80\%)  & 200 (10\%) & 200 (10\%) \\

\hline 
\end{tabular}
\label{table:filted_dataset} 
\end{table}

\paragraph{Mesh pre-processing.} The meshes in the above dataset were pre-processed to (a) \emph{detect and remove interior structure} for each building
(since we aimed to gather annotations of building exteriors), (b) \emph{detect exact duplicates of subgroups} useful for label propagation, as discussed in Section 3.1 (interface for labeling) in our main paper. To detect whether a subgroup is interior, we sample $10$ points per each triangle in the subgroup and shoot rays to $50$ external viewpoints from all these sample points. If a single ray escapes from the subgroup, it is marked as external, otherwise it is internal. We remove all subgroups marked as internal. 
For duplicate detection, we process all-pairs of subgroups in a building. Specifically, for each pair of subgroups, we exhaustively search for upright axis rotations minimizing Chamfer distance. The optimal translation is computed from the difference of the vertex location barycenters. After factoring out the rigid transformation, we compute one-to-one vertex correspondences based on closest pairs in Euclidean space. If  all closest pairs have distance less than $10^{-6}$ of the average OBB diagonals of the subgroups,  we also check if their mesh connectivity matches i.e., the subgroup mesh adjacency matrix is the same given the corresponding vertices. If they match,  the pair is marked as duplicate. Finally, all such pairs are merged into sets containing  subgroups found to be duplicates of each other. 
\vspace{2mm}

\section*{Appendix B: Part labels}
\vspace{1mm}
To determine a set of \emph{common} labels used to identify parts in buildings, we created a variant of our UI that
asked
users to explicitly type tags for selected components instead of selecting labels from a predefined list. We gathered tags from people who have  domain expertise in  the fields of building construction or design. Specifically, we asked $10$ graduate students 
in   civil engineering and architecture
to tag components in a set of $100$ buildings
 uniformly distributed across the different categories. Each student labeled $3$-$10$ different buildings.  We selected tags that appeared at least in $0.5\%$ of the labeled components to filter out uncommon tags. We concatenated the remaining tags with the most frequent tags appearing in the COLLADA leaf nodes  (appearing at least in $0.5\%$ of subgroups). We merged  synonyms and similar tags. 

The resulting list had $39$ tags. During the main phase of annotation of our 2K buildings, $8$ tags were used very sparsely:\ less than $0.05\%$ subgroups throughout the dataset were annotated with these tags: ``ramp'', ``canopy'', ``tympanum'', ``crepidoma'', ``entablature'', ``pediment'', ``bridge'',  and ``deck''. 
We decided them to exclude them from our dataset since the number of train or test subgroups with these labels would be too low (less than $10$, or they existed in only one building).  Any subgroups annotated with these tags were considered as ``unlabeled'' (undetermined) ones.

\begin{table}[t!]
\small
\caption{Statistics regarding mesh resolution in our dataset. From left to right: building category,  
average/median number of faces and vertices.}
\vspace{-4mm}
\centering 
\begin{tabular}{|c||c|c|c|c|}
\hline
\multirow{2}{*}{Category} &
\,avg. \#\, & \,med. \#\, & \,avg. \#\, & \,med. \#\,  \\
& faces & faces & \,vertices\, & \,vertices\, \\  [0.5ex]
\hline 
\hline 
Residential  & 58,522.7 & 32,295.5 & 18,830.6 & 10,684.0 \\
Commercial   &  49,248.5 & 28,862.0 & 16,722.6 & 10,041.0 \\
Religious    & 51,882.7 & 25,979.0 & 16,687.4 & 8,654.0 \\
Civic        & 40,380.1 & 20,512.0 & 13,910.2 & 7,281.0 \\
Castles      & 70,731.2 & 26,493.0 & 21,050.0 & 8,822.0 \\
\hline
Whole Set & 56,250.4  & 29,741.5 & 18,120.9 & 9,845.0 \\

\hline 
\end{tabular}
\label{table:stats} 
\end{table}

\vspace{2mm}
\section*{Appendix C:  Additional dataset statistics}

\vspace{1mm}
As discussed in our main paper, we gathered $10,000$ annotations from qualified MTurkers for $2,000$ buildings ($5$ annotations per building).  Table \ref{table:stats} shows statistics on the polygon resolution of the meshes in our 2K dataset. Table \ref{tab:consistency} reports the worker consistency per part label, which is measured as the  percentage of times that a subgroup label selected by a qualified MTurker agrees wit the majority. Table \ref{table:consistency2} reports the  worker consistency per building category for the training, hold-out validation, and test split. We observe that the worker consistency remains similar across all splits and building categories. 

Table \ref{table:basic_stats} reports statistics on the number of subgroups per building category, unique subgroups (counting repeated subgroups with exactly the same mesh geometry as one unique subgroup), and number of annotated subgroups. We note that there were often subgroups that represented tiny, obscure pieces (e.g., subgroups with a few triangles covering a tiny area of a wall, beam, or frame), and these were often not labeled by annotators. As we explained in the main paper, most of the buildings had more than $80\%$ of their area labeled (and all had $>50\%$ labeled area).
Table \ref{table:more_basic_stats} presents more statistics on the labeled components (merged, adjacent subgroups with the same label) of the $2K$ building dataset per each basic category.

\begin{table}[t!]
\small
\caption{Worker consistency for each different part label. } 
\vspace{-4mm}
\centering 
\renewcommand{\arraystretch}{0.85}
\begin{tabular}{|c|c|}
\hline 
Label & Worker consistency \\  [0.5ex]
\hline 
\hline 
Window        & 93.4\% \\ 
Plant        & 98.7\% \\
Wall        & 88.2\%  \\
Roof        & 88.7\% \\
Banister     & 86.5\% \\
Vehicle        & 99.2\% \\
Door        & 84.9\% \\
Fence        & 85.9\% \\
Furniture     & 95.1\% \\
Column        & 87.2\% \\
Beam        & 76.3\% \\
Tower        & 81.4\% \\
Stairs        & 92.0\% \\
Shutters        & 79.3\% \\
Ground        & 84.8\% \\
Garage        & 86.9\% \\
Parapet        & 82.6\% \\
Balcony        &  75.9\% \\
Floor        & 79.1\% \\
Buttress        & 85.0\% \\
Dome        & 83.5\% \\
Corridor        & 70.6\% \\
Ceiling        & 78.4\% \\
Chimney        & 93.3\% \\
Gate        & 90.8\% \\
Lighting        & 90.9\% \\
Dormer       & 70.4\% \\
Pool        & 86.8\% \\
Road        & 73.8\% \\
Arch        & 72.1\% \\
Awning        & 59.5\%  \\
\hline 
\end{tabular}
\label{tab:consistency} 
\vspace{-1mm}
\end{table}

\begin{table}[t!]
\small
\caption{Worker consistency in the training, hold-out validation, test split, and our whole dataset per  category. } 
\vspace{-4mm}
\centering 
\renewcommand{\arraystretch}{0.85}
\begin{tabular}{|c|c|c|c|c|}
\hline

  & \multicolumn{4}{|c|}{Worker Consistency} \\
  \cline{2-5}
Category  & train. & val. & test & all \\
\hline
\hline
Residential        & 92.2\% & 93.3\%  & 91.5\% & 92.2\% \\
Commercial         & 87.7\% & 89.4\%  & 95.1\% & 88.6\% \\
Religious          & 91.4\% & 91.7\%  & 91.7\% & 91.5\% \\
Civic             & 93.6\% & 98.8\%  & 98.0\% & 94.9\% \\
Castles            & 94.1\% & 88.8\%  & 88.3\% & 92.9\% \\
\hline
\textbf{Average:}  & \textbf{91.8\%} & \textbf{92.4\%}  & \textbf{92.9\%} & \textbf{92.0\%} \\

\hline 
\end{tabular}
\label{table:consistency2} 
\vspace{-1mm}
\end{table}

\begin{table*}[t!]
\small
\caption{Statistics for each building category. From left to right: building category, total number of models, average/median/minimum/maximum number of mesh subgroups over the category's models  (leaf nodes of the  COLLADA metadata of the building models), 
average/median/minimum/maximum number of unique (non-duplicate) subgroups, average/median/minimum/maximum number of annotated unique mesh subgroups.}
\vspace{-4mm}
\centering 
\begin{tabular}{|@{}c@{}||@{}c@{} |@{}c@{} |@{}c@{} |@{}c@{} |@{}c@{} |@{}c@{} |@{}c@{} |@{}c@{} |@{}c@{} |@{}c@{}| @{}c@{}| @{}c@{}| @{}c@{}| @{}c@{}| @{}c@{}| @{}c@{}| @{}c@{}| @{}c@{}| @{}c@{}| @{}c@{}|}
\hline
\multirow{2}{*}{Category} &
\,num\#\, &
\,avg\#\, & \,med\#\, & \,min\#\, & \,max\#\, &
\,avg\# un.\, & \,med\# un.\, & \,min\# un.\, & \,max\# un.\, &
\,avg\# un.\, & \,med\# un.\, & \,min\# un.\, & \,max\# un.\, \\
& \,models\,
& \,subgrps\, & \,subgrps\, & \,subgrps\, & \,subgrps\, &
  \,subgrps\, & \,subgrps\, & \,subgrps\, & \,subgrps\, &
  \,l.subgrps\, & \,l.subgrps\, & \,l.subgrps\, & \,l.subgrps\, \\ 
\hline 
\hline 
Residential  & 1,424 & 678.7 & 547 & 83 & 1989 & 167.1 & 144 & 61 & 920 & 61.4 & 50.0 & 7  & 613  \\
\,Commercial\,  & 153 & 723.4 & 606 & 90 & 1981 & 159.8 & 139 & 70 & 907 & 49.4 & 44.0 & 3  & 223 \\
Religious  & 540 & 487.0 & 348 & 93 & 1981 & 139.9 & 129 & 65 & 667 & 47.2 & 45.0 & 7  & 139 \\
Civic      & 67 & 628.8 & 480 & 118 & 1822 & 144.4 & 123 & 75 & 618 & 43.0 & 43.0 & 8  & 106  \\
Castles & 85 & 609.8 & 485 & 125 & 1786 & 193.0 & 166 & 76 & 590 & 38.6 & 37 & 2  & 92 \\
\hline
Whole Set & 2,000 &\ 623.6 & 497.5 & 83 & 1989 & 160.5 & 140 & 61 & 920 & 55.9 & 47.0 & 2 & 613  \\
\hline 
\end{tabular}
\label{table:basic_stats} 
\end{table*}
  
 \vspace{3mm}
\begin{table*}[t!]
\small
\caption{Statistics per building category regarding components (merged adjacent mesh subgroups).  From left to right: building category, total number of models, 
average/median/minimum/maximum number of annotated components per model, average/median/minimum/maximum number of annotated unique (non-duplicate)  components per model.}
\vspace{-4mm}
\centering 
\begin{tabular}{|@{}c@{}||@{}c@{} |@{}c@{} |@{}c@{} |@{}c@{} |@{}c@{} |@{}c@{} |@{}c@{} |@{}c@{} |@{}c@{} |@{}c@{}| @{}c@{}| @{}c@{}| @{}c@{}| @{}c@{}| @{}c@{}| @{}c@{}|}
\hline
\multirow{2}{*}{Category} &
\,num\#\, &
\,avg\# \, & \,med\# \, & \,min\# \, & \,max\# \, &
\,avg\# un.\, & \,med\# un.\, & \,min\# un.\, & \,max\# un.\, \\
& \,models\, &
  \,l.comp.\, & \,l.comp.\, & \,l.comp.\, & \,l.comp.\, &
  \,l.comp.\, & \,l.comp.\, & \,l.comp.\, & \,l.comp.\, \\ 
\hline 
\hline 
Residential  & 1,424 & 321.8 & 243.0 & 13& 1970& 46.1& 42.0 & 8 & 371 &  \\
\,Commercial\,  & 153 & 408.0 & 296.0& 4& 1680& 44.6 & 39.0 & 3 & 247  \\
Religious  & 540 & 272.2& 184.0& 18& 1469& 37.7 & 35.0 & 6 & 135   \\
Civic      & 67 & 378.4 & 263.0& 36& 1667& 39.3 & 33.0 & 7 & 252 \\
Castles & 85 & 295.3& 210.0 & 40& 1200& 30.5 & 28.0 & 2 & 107  \\
\hline
Whole Set & 2,000 &\ 316.6 & 231.0& 4& 1970 & 43.2 & 39.0 & 2 &\ 371\\
\hline 
\end{tabular}
\label{table:more_basic_stats} 
\end{table*}

\vspace{2mm}
\section*{Appendix D: Network and experiments details}

\vspace{2mm}
\paragraph{BuildingGNN.} We provide more details about the structure of the BuildingGNN network architecture in Table \ref{table:layers}.
Table \ref{table:gnnedge_stats} presents statistics on the number of  edges per type used in BuildingGNN for our training set.

\begin{table}[t!]
\centering 
\caption{BuildingGNN architecture: The Node representation combines the OBB - (Object Oriented Bounding Box) , SA - (Surface area), C - (centroid) and MN - (MinkowskiNet pre-trained features) for each sub group. The GNN is composed of (a) an encoder block made of three MLPs having 1, 3 and 5 hidden layers respectively, and (b) a decoder block with one MLP having 1 hidden layer followed by softmax. We refer to the code for more details.}
\vspace{-4mm}
\small
\begin{tabular}{|@{}c@{}|@{}c@{}|@{}c@{}|}
\hline
\hline

    & Layers                                            & Output   \\
    \hline
    \hline
    \multirow{1}{*}{Edge}  & (MLP(11$\times$41, layer=1)))  &$41$\\\cline{2-3}
    \hline
    \multirow{1}{*}{Node}& (6D(OBB)+1D(SA)+3D(C)+31D(MN) & $41$\\\cline{2-3}
    \hline
    \multirow{1}{*}{Input} & ($\text{Node}_i$ + $\text{Edge}_{ij}$ + $\text{Node}_j$) & $41$ \\\cline{2-3}
\hline
\hline
\multirow{3}{*}{Encoder}  & (MLP(Input$\times$256, layer=1)))  &$64$\\\cline{2-3}
                          & GN(LeakyReLU(0.2)))  &$64$\\\cline{2-3}
                          & (MLP(64*3$\times$128, layer=3)))  &$64$\\\cline{2-3}
                          & GN(LeakyReLU(0.2)))  &$64$\\\cline{2-3}
                          & (MLP(64*3$\times$128, layer=5)))  &$64$\\\cline{2-3}
                          & GN(LeakyReLU(0.2)))  &$64$\\\cline{2-3}
                         
\hline
\hline
\multirow{2}{*}{Decoder}  & (MLP(128$\times$64, layer=1)))  &$31$\\\cline{2-3}
 & softmax  & $31$\\\cline{2-3}
\hline
\hline 
\end{tabular}
\label{table:layers} 
\end{table}

\begin{center}
\begin{table}[t!]
\begin{center}
\caption{Statistics for the number of BuildingGNN edges per type present in the graphs of the training buildings. }
\vspace{-4mm}
\begin{tabular}[t]{|c|c|c|c|c|} 
\hline
 \multirow{2}{*}{Label} & max \#  & min \#  & mean \# & \# median \\
  &  \,edges\,    &  edges   & edges  & edges  \\  [0.5ex]
 \hline \hline
 Proximity & 16317 &      81   & 778.0  & 489.0   \\ 
 \hline
 Similarity & 762156 & 5   & 26452.1   & 4875.5  \\
 \hline
 Containment & 26354     & 71   & 2,054.5   & 1,390.0  \\
  \hline
 Support & 7234     & 7   & 687.5   & 492.0   \\
  \hline
 All & 772878     & 259   & 29972.1  & 7818.0  \\
 \hline
\end{tabular}
\label{table:gnnedge_stats}
\end{center}
\end{table}
\end{center}

\vspace{-1mm}
\paragraph{MinkNet-GC.} As mentioned in the experiments section of our main paper, we implemented a simple graph-cuts variant, called MinkNet-GC, that incorporates label probabilities from MinkowskiUNet34 as unary terms, and a pairwise term that depends  on angles between triangles, inspired by \cite{Kalogerakis10}. Specifically, we use the following energy that we minimize using
\cite{boykov}:

\begin{equation}
    E(\pmb{y}) = \sum_{i \in \mathcal{F}} \psi(y_i) + \sum_{i \in \mathcal{F}} \sum_{j \in \mathcal{N}(i)} \phi(y_i, y_j)
\label{eq:graph_cuts}
\end{equation}
where $\pmb{y}=\{y_i\}$ are the label assignments we wish to compute by minimizing the above energy, $\mathcal{F}$ is the set of faces in a mesh, and $\mathcal{N}(i)$ are the adjacent faces of each face $i$. The unary term is expressed as follows: 
\mbox{$\psi(y_i)=-\log f(y_i)$}, where $f(y_i)$ is the probability distribution over part labels associated with the face $i$ produced through average pooling of probabilities computed from MinkowskiUNet34 on the triangle's associated points. The 
pairwise term uses angles between face normals, \mbox{$\phi'(y_i, y_j)=-\lambda' \cdot \log ( \min(\omega_{i,j}/ 90^o,1) )$}, for \mbox{$y_i \ne y_j$},  where $\omega_{i,j}$ is the angle between the normals of faces $i,j$. The term results in zero cost for right angles between 
normals indicating a strong edge.  The  parameter $\lambda$ is adjusted with grid search in the hold-out validation set. 

\vspace{-1mm}
\paragraph{Average vs max pooling.} As discussed in our experiments section of our main paper, one possibility to aggregate probabilities of points associated per triangle or component is average pooling:
 $\bq_t = \sum_{p \in P_t} \bq_p / |P_t| $ where $\bq_p$ and $\bq_t$ are point and triangle probabilities respectively. An alternative is to use max pooling (i.e., replace sum with max above). We experimented with average vs max pooling also per component. As shown in Table \ref{table:mesh_track_max_pooling}, average pooling works better for both triangle- and component-based pooling (we experimented with MinkowskiNet per-point probabilities).

\begin{table}[t!]
\small
\caption{``BuildingNet-Mesh'' results using average and max pooling aggregation over triangles and components (weighted cross-entropy loss was used for all these experiments).} 
\vspace{-4mm}
\centering 
\renewcommand{\arraystretch}{0.85}
\begin{tabular}{|@{}c@{}|@{}c@{}|@{}c@{}|@{}c@{}|@{}c@{}|@{}c@{}|@{}c@{}|}
\hline
Method  &  Pool.& \,$\bn$?\, & \,$\bc$?\, & Part IoU  & Shape IoU & Class acc. \\
\hline 
\hline 
\multirow{4}{*}{MinkNet2Triangle} & Avg & $\checkmark$ & $\times$ & \textbf{28.8\%} & \textbf{26.7\%} & \textbf{64.8\%}\\
                                  & Max & $\checkmark$ & $\times$ & 28.6\% & 26.1\% & 64.4\%\\
                                  \cline{2-7}
                                  & Avg & $\checkmark$ & $\checkmark$ & \textbf{32.8\%} & \textbf{29.2\%} & \textbf{68.1\%}\\
                                  & Max & $\checkmark$ & $\checkmark$ & 31.5\% & 28.1\% & 66.8\%\\
\hline
\hline
\multirow{4}{*}{MinkNet2Sub} & Avg & $\checkmark$ & $\times$ & \textbf{33.1\%} & \textbf{36.0\%} & \textbf{69.9\%}\\
                             & Max & $\checkmark$ & $\times$ & 30.4\% & 32.4\% & 65.6\%\\
                             \cline{2-7}
                             & Avg & $\checkmark$ & $\checkmark$ & \textbf{37.0\%} & \textbf{39.1\%} & \textbf{73.2\%}\\
                             & Max & $\checkmark$ & $\checkmark$ & 32.7\% & 34.8\% & 67.4\%\\
\hline 
\end{tabular}
\label{table:mesh_track_max_pooling} 
\end{table}

\vspace{-1mm}
\paragraph{Experiments with different losses.} We experimented with different losses for our MinkowskiNet variants for the ``BuildingNet-Point''  and ``BuildingNet-Mesh'' tracks. Specifically, we experimented with  the Weighted Cross-Entropy Loss (WCE) described in our main paper, Cross-Entropy Loss (CE) without label weights, the Focal Loss (FL) \cite{focal}, $\alpha$-balanced Focal Loss ($\alpha$-FL) \cite{focal}, and  Class-Balanced Cross Entropy Loss (CB) \cite{cui2019}. Table \ref{table:point_cloud_track_losses_ablation} and Table \ref{table:mesh_track_losses_ablation} show results for the ``BuildingNet-Point''  and ``BuildingNet-Mesh'' tracks respectively. We observe that (a) in the case that color is not available, WCE is slightly better than alternatives according to all measures for both tracks (b) when color is available, CB is a bit better in terms of Part IoU, but worse in terms of Shape IoU than WCE in the case of the point cloud track. For the mesh track, CB is slightly better according to all measures. In general, WCE and CB behave the best on average, yet their difference is small. For the rest of our experiments, we use WCE.

\vspace{-1mm}
\paragraph{Performance for each part label.} Our main paper reports mean Part IoU performance in the experiments section. Table \ref{table:IOU} reports the BuildingGNN-PointNet++ and BuildingGNN-MinkNet part IoU performance for each label. We also report the performance of MinkowskiNet and PointNet++ for the point cloud track. We observe that networks do better for common part labels, such as window, wall, roof, plant, vehicle, while the performance degrades  for rare parts (e.g., awning, arch), or parts whose shape can easily be confused with other more dominant parts (e.g., garage is often confused with door, wall, or window).

\begin{table}[h!]
\small
\caption{``BuildingNet-Point'' track results using the Weighted Cross-Entropy Loss (WCE), Cross-Entropy Loss (CE), Focal Loss (FL), $\alpha$-balanced Focal Loss ($\alpha$-FL) and finally Class-Balanced Cross Entropy Loss (CB). All these were used to train the MinkowskiUNet34 architecture. For the FL and $\alpha$-FL experiments the $\gamma$ hyper-parameter was set to $2.0$ and for the $\alpha$-FL the same weights were used as the weighted cross entropy loss (see Section 4.3 in our main paper). For the CB experiments we set $\beta=0.999999$.} 
\vspace{-3mm}
\centering
\renewcommand{\arraystretch}{0.85}
\begin{tabular}{|c|c|@{}c@{}|@{}c@{}|c|c|c|}
\hline
Method  &  Loss & \,$\bn$?\, & \,$\bc$?\, & Part IoU  & Shape IoU & Class acc. \\
\hline 
\hline 
\multirow{5}{*}{MinkNet} & WCE & $\checkmark$ & $\times$ & \textbf{26.9\%} & \textbf{22.2\%} & \textbf{62.2\%}\\
                         & CE & $\checkmark$ & $\times$ & 24.5\% & 21.2\% & 61.3\%\\
                         & FL & $\checkmark$ & $\times$ & 26.1\% & 21.8\% & 61.2\%\\
                         & $\alpha$-FL & $\checkmark$ & $\times$ & 22.3\% & 19.8\% & 61.5\%\\
                         & CB & $\checkmark$ & $\times$ & 26.4\% & 20.9\% & 61.4\%\\
\hline
\hline
\multirow{5}{*}{MinkNet} & WCE & $\checkmark$ & $\checkmark$ & 29.9\% & 24.3\% & \textbf{65.5\%}\\
                         & CE & $\checkmark$ & $\checkmark$ & 28.5\% & 24.5\% & 65.3\%\\
                         & FL & $\checkmark$ & $\checkmark$ & 28.7\% & 24.9\% & 65.2\%\\
                         & $\alpha$-FL & $\checkmark$ & $\checkmark$ & 30.1\% & \textbf{25.3\%} & 65.2\%\\
                         & CB & $\checkmark$ & $\checkmark$ & \textbf{30.4\%} & 24.0\% & \textbf{65.5\%}\\
\hline 
\end{tabular}
\label{table:point_cloud_track_losses_ablation} 
\end{table}

\begin{table}[h!]
\small
\caption{``BuildingNet-Mesh'' results using different loss functions} 
\vspace{-4mm}
\centering
\renewcommand{\arraystretch}{0.85}
\begin{tabular}{|@{}c@{}|@{}c@{}|@{}c@{}|@{}c@{}|@{}c@{}|@{}c@{}|@{}c@{}|}
\hline
Method  &  Loss & \,$\bn$?\, & \,$\bc$?\, & Part IoU  & Shape IoU & Class acc. \\
\hline
\hline
\multirow{5}{*}{MinkNet2Sub} & WCE & $\checkmark$ & $\times$ & \textbf{33.1\%} & \textbf{36.0\%} &         \textbf{69.9\%}\\
                         & CE & $\checkmark$ & $\times$ & 30.7\% & 32.7\% & 68.8\%\\
                         & FL & $\checkmark$ & $\times$ & 31.0\% & 33.4\% & 67.9\%\\
                         & $\alpha$-FL & $\checkmark$ & $\times$ & 27.2\% & 28.3\% & 66.7\%\\
                         & CB & $\checkmark$ & $\times$ & 32.9\% & 34.3\% & 69.1\%\\
\hline
\hline
\multirow{5}{*}{MinkNet2Sub} & WCE & $\checkmark$ & $\checkmark$ & 37.0\% & 39.1\% & 73.2\%\\
                         & CE & $\checkmark$ & $\checkmark$ & 35.6\% & 39.2\% & 73.5\%\\
                         & FL & $\checkmark$ & $\checkmark$ & 35.1\% & 38.4\% & 73.2\%\\
                         & $\alpha$-FL & $\checkmark$ & $\checkmark$ & 36.0\% & 38.2\% & 72.4\%\\
                         & CB & $\checkmark$ & $\checkmark$ & \textbf{38.0\%} & \textbf{39.7\%} & \textbf{73.9\%}\\
\hline 
\end{tabular}
\label{table:mesh_track_losses_ablation} 
\end{table}

\begin{table}[h!]
\small
\caption{BuildingGNN ablation study  based on PointNet++ node features}
\vspace{-4mm}
\centering 
\begin{tabular}{|@{}c@{}|@{}c@{}|@{}c@{}|@{}c@{}|@{}c@{}|@{}c@{}|}
\hline
Variant  &  \,$\bn$?\, & \,$\bc$?\, & \,Part IoU\,  & \,Shape IoU\, &\, Class acc. \\
\hline 
\hline
Node-OBB     &  $\checkmark$  & $\checkmark$ & 10.0\%  & 17.1\% & 56.5\%\\
Node-PointNet++    &  $\checkmark$  & $\checkmark$  & 14.0\%  & 19.1\% & 52.2\%\\
Node-OBB+PointNet++ &  $\checkmark$  & $\checkmark$  & 24.4\%  & 27.8\% & 71.7\%\\
 w/ support edges &  $\checkmark$  & $\checkmark$  & 26.7\%  & 29.2\% & 71.5\%\\
w/ containment edges &  $\checkmark$  & $\checkmark$  & 27.9\%  &30.6\% & 72.6\%\\
w/ proximity edges &  $\checkmark$  & $\checkmark$  & 26.4\%  & 29.4\% & 71.4\%\\
w/ similarity edges &  $\checkmark$  & $\checkmark$  & 23.1\%  & 28.5\% & 69.8\%\\
BuildingGNN-PointNet++ &  $\checkmark$  & $\checkmark$  & \textbf{31.5\%}  & \textbf{35.9\%} & \textbf{73.9\%} \\
\hline
\end{tabular}
\label{table:ablation_pointnet} 
\end{table}

\begin{table}[h!]
\small
\caption{BuildingGNN ablation study based on MinkowskiNet node features}
\vspace{-4mm}
\centering 
\begin{tabular}{|@{}c@{}|@{}c@{}|@{}c@{}|@{}c@{}|@{}c@{}|@{}c@{}|@{}c@{}|}
\hline
Variant  &  \,$\bn$?\, & \,$\bc$?\, & \,Part IoU\,  & \,Shape IoU\, &\, Class acc. \\
\hline 
\hline
Node-OBB     &  $\checkmark$  & $\checkmark$ & 10.0\%  & 17.1\% & 56.5\%\\
Node-MinkNet    &  $\checkmark$  & $\checkmark$  & 35.6\%  & 35.9\% & 67.7\%\\
Node-OBB+MinkNet &  $\checkmark$  & $\checkmark$  & 40.0\%  & 40.6\% & 75.8\%\\
 w/ support edges &  $\checkmark$  & $\checkmark$  & 42.0\%  & 43.5\% & 77.8\%\\
w/ containment edges &  $\checkmark$  & $\checkmark$  & 41.1\%  & 42.0\% & 76.8\%\\
w/ proximity edges &  $\checkmark$  & $\checkmark$  & 39.9\%  & 40.6\% & 75.6\%\\
w/ similarity edges &  $\checkmark$  & $\checkmark$  & 41.2\%  & 43.0\% & 75.8\%\\
BuildingGNN-MinkNet &  $\checkmark$  & $\checkmark$  & \textbf{42.6\%}  & \textbf{46.8\%} &\textbf{77.8\%} \\
\hline

\end{tabular}
\label{table:ablation_mink} 
\end{table}

\begin{table*}[t!]
\small
\caption{ Part IOU performance  for each label. BuildingGNN-MinkNet and BuildingGNN-PointNet++ are tested on the mesh track, while MinkNet and PointNet++ are tested on the point cloud track. The left half of the table reports performance when color is available (``n+c''), while the right half reports performance when it is not available (``n'').} 
\vspace{-10mm}
\renewcommand{\arraystretch}{0.85}
\centering
\begin{center}
\begin{tabular}{@{}c@{}|@{}c@{}}
\hline
\end{tabular}
\end{center}
\begin{tabular}{|@{}c@{}|@{}c@{}|@{}c@{}|@{}c@{}|@{}c@{}||@{}c@{}|@{}c@{}|@{}c@{}|@{}c@{}|}
\hline 
\hline
 \multirow{2}{*}{Label} & \, BuildingGNN \,&  \,BuildingGNN\,& \, MinkNet\,  & \,PointNet++ \,&\, BuildingGNN \,&  \,BuildingGNN\,& \, MinkNet\,  & \,PointNet++ \,\\
  &  \,MinkNet(n+c)\,    &  PointNet++(n+c)   &(n+c) &(n+c) & \,MinkNet(n)\, &\, PointNet++(n) & (n) & (n) \\  [0.5ex]
\hline 
\hline 

Window        & 70.5\% & 71.1\% &44.1\% &34.8\% & 70.4\% &68.3\% & 35.6\% &0.0\% \\
Plant        & 81.0\% &69.8\% &79.6\% &70.3\% & 79.8\% &69.8\% & 79.7\% &48.4\% \\
Vehicle        & 83.7\% & 77.3\%& 77.1\%& 29.7\% & 82.7\% &72.4\% & 75.8\% &19.2\% \\
Wall        & 78.1\% &77.5\% &64.5\% &57.9\% & 76.0\% &74.4\% & 63.2\% &54.4\% \\
Banister        & 50.0\% & 19.9\%& 44.9\%& 0.0\% & 56.5\% &22.0\% & 45.6\% &0.0\% \\
Furniture       & 59.7\% &37.0\% & 56.0\%& 0.0\% & 58.3\% &43.5\% & 54.9\% &0.0\% \\
Fence        & 55.5\% &34.7\% & 71.3\%& 16.5\% & 64.1\% &19.7\% & 49.5\% &9.6\% \\
Roof        & 78.9\% &72.1\% & 65.3\%& 58.2\% & 70.2\% &69.0\% & 67.0\% &56.4\% \\
Door        & 41.7\% &37.6\% & 21.7\%& 0.0\% & 39.2\% &37.7\% & 23.8\% &0.0\% \\
Tower        & 53.4\% &41.2\% & 46.5\%& 2.3\% & 50.8\% &37.5\% & 43.4\% &4.8\% \\
Column        & 61.5\% &27.6\% & 49.5\%& 0.6\% & 53.6\% &34.7\% & 42.9\% &1.1\% \\
Beam        & 24.9\% & 22.4\% & 13.8\%& 0.02\% & 30.3\% &21.5\% & 17.2\% &0.0\% \\
Stairs        & 38.6\% &25.6\% & 26.9\%&0.0\% & 41.0\% &24.1\% & 27.8\% &0.0\% \\
Shutters        & 1.0\% &1.3\% & 0.0\%& 0.0\% & 1.7\% &0.0\% & 0.0\% &0.0\% \\
Garage        & 9.0\% &10.6\% & 3.6\%& 0.0\% & 10.6\% &8.4\% & 6.8\% &0.0\% \\
Parapet        & 24.9\% &3.9\% & 11.6\%&0.0\% & 28.6\% &2.5\% & 21.0\% &0.0\% \\
Gate        & 14.0\% & 16.5\%&6.4\% &0.0\% & 7.9\% &12.3\% & 7.9\% &0.0\% \\
Dome        & 53.8\% &10.1\% & 48.0\% & 1.9\% & 54.3\% &14.2\% & 54.5\% &16.3\% \\
Floor       & 51.5\% &37.7\% & 47.8\%& 36.9\% & 51.2\% &30.9\% & 46.8\% &30.0\% \\
Ground        & 75.0\% &65.1\% & 77.4\% & 64.1\% & 61.8\% &55.5\% & 60.8\% &42.6\% \\
Buttress        & 23.8\% &9.6\% & 15.6\% & 0.0\% & 38.7\% &12.3\% & 6.1\% &0.0\% \\
Balcony        & 19.6\% &9.5\% & 15.0\% & 0.0\% & 15.5\% &15.6\% & 17.3\% &0.0\% \\
Chimney        & 70.0\% &50.9\% & 57.9\% & 0.0\% & 53.6\% &49.5\% & 60.1\% &0.0\% \\
Lighting        & 6.4\% &9.1\% & 16.8\%& 0.0\% & 24.9\% &3.5\% & 23.3\% &0.0\% \\
Corridor        & 16.3\% &10.5\% & 15.9\%& 4.2\% & 7.2\% &4.1\% & 7.2\% &0.0\% \\
Ceiling        & 28.0\% &23.8\% &22.1\% & 4.6\% & 28.0\% &20.5\% & 17.4\% &4.6\% \\
Pool        & 70.8\% & 53.0\%& 78.7\% & 77.8\% & 38.1\% &33.0\% & 43.0\% &0.0\% \\
Dormer        & 27.3\% & 20.4\% & 9.6\%& 0.0\% & 22.1\% &23.3\% & 6.8\% &0.0\% \\
Road        & 46.2\% &24.1\% & 53.5\% & 40.0\% & 1.9\% &16.3\% & 21.5\% &0.0\% \\
Arch        & 8.4\% & 5.2\%& 0.9\% & 0.0\% & 3.2\% &2.9\% & 0.8\% &0.0\% \\
Awning        & 1.5\% &0\% & 3.8\%& 0.0\% & 1.6\% &0.0\% & 0.0\% &0.0\% \\

\hline 
\end{tabular}
\label{table:IOU} 
\end{table*}

\section*{Appendix E: BuildingGNN ablation study}

We conducted an ablation study involving different node features, and also experimenting with different types of edges in our BuildingGNN. Table \ref{table:ablation_pointnet} present the results for different experimental conditions of our BuildingGNN based on PointNet++ as node features. We first experimented using no edges and processing node features alone through our MLP structure. We experimented with using only OBB-based features (``Node-OBB''), using features from PointNet++ alone (``Node-PointNet++''), and finally using both node features concatenated (``Node-OBB+PointNet++''). 
We observe that  using all combinations of node features yields better performance compared to using either node feature type alone. Then we started experimented with adding each type of edges individually to our network (e.g., ``w/ support edges'' in Table \ref{table:ablation_pointnet}  means that we use node features with support edges only). Adding each type of edge individually further boosts performance compared to using node features alone. Using all edges (``BuildingGNN-PointNet'') yields a noticeable $7.1\%$ Part IoU increase and $8.1\%$ Shape IoU increase compared to using node features alone.
Table \ref{table:ablation_mink} shows the same experiments using MinkowskiNet-based features. We observe that combined node features perform better than using either node feature type alone. Adding each type of edges helps, except for proximity edges that seem to have no improvement when used alone. Using all edges still yields a noticeable $2.6\%$ Part IoU increase and $6.2\%$ Shape IoU increase compared to using node features alone. 

We also experimented with DGCNN \cite{dgcnn} as a backbone in our GNN for extracting node features.  Unfortunately, DGCNN could not directly handle our large points clouds (100K points). It runs out of memory even with batch size $1$ on a 48GB GPU card. We tried to downsample the point clouds (10K points) to pass them to DGCNN, then propagated the node features back to the $100$K points using nearest neighbor upsampling. The part IoU was $32.5\%$ in the mesh track with color input and using all edges (i.e., the performance is comparable to BuildingGNN-PointNet++, but much lower than  BuildingGNN-MinkNet). Still, since other methods were able to handle the original resolution without downsampling, this comparison is not necessarily fair, thus we excluded it from the tables showing the track results in our main paper.

\vspace{2mm}
\section*{Appendix F: Additional Material}
\vspace{1mm}
We refer readers to our project page
\mbox{\url{www.buildingnet.org}} for the dataset, source code, and other supplementary material.

%% file: BuildingNet.bbl
\begin{thebibliography}{10}\itemsep=-1pt

\bibitem{Armeni:2016}
I. {Armeni}, O. {Sener}, A.~R. {Zamir}, H. {Jiang}, I. {Brilakis}, M.
  {Fischer}, and S. {Savarese}.
\newblock {3D Semantic Parsing of Large-Scale Indoor Spaces}.
\newblock In {\em Proc. CVPR}, 2016.

\bibitem{behley2019iccv}
J. Behley, M. Garbade, A. Milioto, J. Quenzel, S. Behnke, C. Stachniss, and J.
  Gall.
\newblock {SemanticKITTI: A Dataset for Semantic Scene Understanding of LiDAR
  Sequences}.
\newblock In {\em Proc. ICCV}, 2019.

\bibitem{Boscaini2016}
Davide Boscaini, Jonathan Masci, Emanuele Rodol{\`a}, and Michael Bronstein.
\newblock {Learning shape correspondence with anisotropic convolutional neural
  networks}.
\newblock In {\em Proc. NIPS}, 2016.

\bibitem{boykov}
Yuri Boykov, Olga Veksler, and Ramin Zabih.
\newblock Fast approximate energy minimization via graph cuts.
\newblock {\em IEEE Trans. Pat. Ana. \& Mach. Int.}, 23(11), 2001.

\bibitem{Chang:2017}
Angel Chang, Angela Dai, Thomas Funkhouser, Maciej Halber, Matthias Niessner,
  Manolis Savva, Shuran Song, Andy Zeng, and Yinda Zhang.
\newblock {Matterport3D: Learning from RGB-D Data in Indoor Environments}.
\newblock In {\em Proc. 3DV}, 2017.

\bibitem{Chang:2015}
Angel~X. Chang, Thomas Funkhouser, Leonidas Guibas, Pat Hanrahan, Qixing Huang,
  Zimo Li, Silvio Savarese, Manolis Savva, Shuran Song, Hao Su, Jianxiong Xiao,
  Li Yi, and Fisher Yu.
\newblock {ShapeNet: An Information-Rich 3D Model Repository}.
\newblock Technical Report arXiv:1512.03012 [cs.GR], Stanford University ---
  Princeton University --- Toyota Technological Institute at Chicago, 2015.

\bibitem{Chen:CGF:2020}
L. Chen, W. Tang, N.~W. John, T.~R. Wan, and J.~J. Zhang.
\newblock Context-aware mixed reality: A learning-based framework for
  semantic-level interaction.
\newblock {\em Computer Graphics Forum}, 39(1), 2020.

\bibitem{Chen:2009}
Xiaobai Chen, Aleksey Golovinskiy, and Thomas Funkhouser.
\newblock {A} {B}enchmark for 3{D} {M}esh {S}egmentation.
\newblock {\em ACM Trans. on Graphics}, 28(3), 2009.

\bibitem{choy20194d}
Christopher Choy, JunYoung Gwak, and Silvio Savarese.
\newblock {4D} {S}patio-{T}emporal {C}onv{N}ets: {M}inkowski {C}onvolutional
  {N}eural {N}etworks.
\newblock In {\em Proc. CVPR}, 2019.

\bibitem{cui2019}
Yin Cui, Menglin Jia, Tsung-Yi Lin, Yang Song, and Serge Belongie.
\newblock {C}lass-{B}alanced {L}oss {B}ased on {E}ffective {N}umber of
  {S}amples.
\newblock In {\em Proc. CVPR}, 2019.

\bibitem{Dai:2017}
Angela Dai, Angel~X. Chang, Manolis Savva, Maciej Halber, Thomas Funkhouser,
  and Matthias Nie{\ss}ner.
\newblock {ScanNet: Richly-annotated 3D Reconstructions of Indoor Scenes}.
\newblock In {\em Proc. CVPR}, 2017.

\bibitem{Demir:2015_2}
Ilke Demir, Daniel~G. Aliaga, and Bedrich Benes.
\newblock {Coupled Segmentation and Similarity Detection for Architectural
  Models}.
\newblock {\em ACM Trans. on Graphics}, 34(4), 2015.

\bibitem{Demir:2015}
I. {Demir}, D.~G. {Aliaga}, and B. {Benes}.
\newblock {Procedural Editing of 3D Building Point Clouds}.
\newblock In {\em Proc. ICCV}, 2015.

\bibitem{fu20203dfuture}
Huan Fu, Rongfei Jia, Lin Gao, Mingming Gong, Binqiang Zhao, Steve Maybank, and
  Dacheng Tao.
\newblock {3D-FUTURE: 3D Furniture shape with TextURE}.
\newblock {\em arXiv preprint arXiv:2009.09633}, 2020.

\bibitem{google2017maps}
{Google Maps}.
\newblock \url{https://maps.google.com}, 2017.

\bibitem{arteta2017history}
Jon~Arteta Grisale$\tilde{\textrm{n}}$a.
\newblock {\em The Paradigm of Complexity in Architectural and Urban Design
  ({PhD} Thesis)}.
\newblock University of Alcala, 2017.

\bibitem{hackel2017isprs}
Timo Hackel, N. Savinov, L. Ladicky, Jan~D. Wegner, K. Schindler, and M.
  Pollefeys.
\newblock {SEMANTIC3D.NET: A new large-scale point cloud classification
  benchmark}.
\newblock In {\em Proc. ISPRS}, 2017.

\bibitem{Hanocka:2019}
Rana Hanocka, Amir Hertz, Noa Fish, Raja Giryes, Shachar Fleishman, and Daniel
  Cohen-Or.
\newblock {MeshCNN}: {A} {N}etwork with an {E}dge.
\newblock {\em ACM Trans. on Graphics}, 38(4), 2019.

\bibitem{Hu:2012}
Ruizhen Hu, Lubin Fan, and Ligang Liu.
\newblock {C}o-{S}egmentation of 3{D} {S}hapes via {S}ubspace {C}lustering.
\newblock {\em Computer Graphics Forum}, 31(5), 2012.

\bibitem{Hua:2016}
Binh-Son Hua, Quang-Hieu Pham, Duc~Thanh Nguyen, Minh-Khoi Tran, Lap-Fai Yu,
  and Sai-Kit Yeung.
\newblock {SceneNN: A Scene Meshes Dataset with aNNotations}.
\newblock In {\em Proc. 3DV}, 2016.

\bibitem{Kalogerakis:2017:ShapePFCN}
Evangelos Kalogerakis, Melinos Averkiou, Subhransu Maji, and Siddhartha
  Chaudhuri.
\newblock {3D} {S}hape {S}egmentation with {P}rojective {C}onvolutional
  {N}etworks.
\newblock In {\em Proc. CVPR}, 2017.

\bibitem{Kalogerakis10}
Evangelos Kalogerakis, Aaron Hertzmann, and Karan Singh.
\newblock {L}earning {3}{D} {M}esh {S}egmentation and {L}abeling.
\newblock {\em ACM Trans. on Graphics}, 29(3), 2010.

\bibitem{Kingma15}
Diederik~P. Kingma and Jimmy Ba.
\newblock {A}dam: {A} {M}ethod for {S}tochastic {O}ptimization.
\newblock In {\em Proc. ICLR}, 2015.

\bibitem{kipf2018neural}
Thomas Kipf, Ethan Fetaya, Kuan-Chieh Wang, Max Welling, and Richard Zemel.
\newblock {N}eural {R}elational {I}nference for {I}nteracting {S}ystems.
\newblock In {\em Proc. ICML}, 2018.

\bibitem{Kobyshev16}
N. {Kobyshev}, H. {Riemenschneider}, A. {Bodis-Szomoru}, and L. {Van Gool}.
\newblock {Architectural decomposition for 3D landmark building understanding}.
\newblock In {\em Proc. WACV}, 2016.

\bibitem{Akundu:2020}
Abhijit Kundu, Xiaoqi Yin, Alireza Fathi, David Ross, Brian Brewington, Thomas
  Funkhouser, and Caroline Pantofaru.
\newblock {Virtual Multi-view Fusion for 3D Semantic Segmentation}.
\newblock In {\em Proc. ECCV}, 2020.

\bibitem{Lahav:2020}
Alon Lahav and Ayellet Tal.
\newblock {MeshWalker: Deep Mesh Understanding by Random Walks}.
\newblock {\em ACM Trans. on Graphics (Proc. SIGGRAPH Asia)}, 39(6), 2020.

\bibitem{Landrieu:2018}
L. {Landrieu} and M. {Simonovsky}.
\newblock {Large-Scale Point Cloud Semantic Segmentation with Superpoint
  Graphs}.
\newblock In {\em Proc. CVPR}, 2018.

\bibitem{InteriorNet18}
Wenbin Li, Sajad Saeedi, John McCormac, Ronald Clark, Dimos Tzoumanikas, Qing
  Ye, Yuzhong Huang, Rui Tang, and Stefan Leutenegger.
\newblock {InteriorNet: Mega-scale Multi-sensor Photo-realistic Indoor Scenes
  Dataset}.
\newblock In {\em Proc. BMVC}, 2018.

\bibitem{focal}
Tsung-Yi Lin, Priya Goyal, Ross~B. Girshick, Kaiming He, and Piotr Doll{\'a}r.
\newblock {F}ocal {L}oss for {D}ense {O}bject {D}etection.
\newblock In {\em Proc. ICCV}, 2017.

\bibitem{loshchilov2017sgdr}
I. Loshchilov and F. Hutter.
\newblock {SGDR}: {S}tochastic {G}radient {D}escent with {W}arm {R}estarts.
\newblock In {\em Proc. ICLR}, 2017.

\bibitem{Lun:2015:StyleSimilarity}
Zhaoliang Lun, Evangelos Kalogerakis, and Alla Sheffer.
\newblock {Elements of Style: Learning Perceptual Shape Style Similarity}.
\newblock {\em ACM Trans. on Graphics}, 34(4), 2015.

\bibitem{mahmud2020overhead}
Jisan Mahmud, True Price, Akash Bapat, and Jan-Michael Frahm.
\newblock Boundary-aware {3D} building reconstruction from a single overhead
  image.
\newblock In {\em Proc. CVPR}, 2020.

\bibitem{masci2015geodesic}
Jonathan Masci, Davide Boscaini, Michael Bronstein, and Pierre Vandergheynst.
\newblock {Geodesic convolutional neural networks on {R}iemannian manifolds}.
\newblock In {\em Proc. ICCV Workshops}, 2015.

\bibitem{Mikolov}
Tomas Mikolov, Ilya Sutskever, Kai Chen, Greg Corrado, and Jeffrey Dean.
\newblock {D}istributed {R}epresentations of {W}ords and {P}hrases and {T}heir
  {C}ompositionality.
\newblock In {\em Proc. NIPS}, 2013.

\bibitem{Mitra2006}
Niloy~J. Mitra, Leonidas~J. Guibas, and Mark Pauly.
\newblock {Partial and Approximate Symmetry Detection for 3D Geometry}.
\newblock {\em ACM Trans. on Graphics}, 25(3), 2006.

\bibitem{Mo_2019_CVPR}
Kaichun Mo, Shilin Zhu, Angel~X. Chang, Li Yi, Subarna Tripathi, Leonidas~J.
  Guibas, and Hao Su.
\newblock {PartNet}: A {L}arge-{S}cale {B}enchmark for {F}ine-{G}rained and
  {H}ierarchical {P}art-{L}evel 3{D} {O}bject {U}nderstanding.
\newblock In {\em Proc. CVPR}, 2019.

\bibitem{Monti2017}
Federico Monti, Davide Boscaini, Jonathan Masci, Emanuele Rodola, Jan Svoboda,
  and Michael~M Bronstein.
\newblock {Geometric deep learning on graphs and manifolds using mixture model
  cnns}.
\newblock In {\em Proc. CVPR}, 2017.

\bibitem{Munoz2009}
D. {Munoz}, J.~A. {Bagnell}, N. {Vandapel}, and M. {Hebert}.
\newblock {Contextual classification with functional Max-Margin Markov
  Networks}.
\newblock In {\em Proc. CVPR}, 2009.

\bibitem{nguyen2016robust}
Duc~Thanh Nguyen, Binh-Son Hua, Lap-Fai Yu, and Sai-Kit Yeung.
\newblock A robust 3{D}-2{D} interactive tool for scene segmentation and
  annotation.
\newblock {\em IEEE Trans. Vis. \& Comp. Graphics}, 24(12), 2018.

\bibitem{qi2017pointnetplusplus}
Charles~R Qi, Li Yi, Hao Su, and Leonidas~J Guibas.
\newblock {PointNet++}: {D}eep {H}ierarchical {F}eature {L}earning on {P}oint
  {S}ets in a {M}etric {S}pace.
\newblock In {\em Proc. NIPS}, 2017.

\bibitem{qiao2019laplaciannet}
Yi-Ling Qiao, Lin Gao, Jie Yang, Paul~L. Rosin, Yu-Kun Lai, and Xilin Chen.
\newblock {Learning on 3D Meshes with Laplacian Encoding and Pooling}.
\newblock {\em IEEE Trans. Vis. \& Comp. Graphics}, 2020.

\bibitem{Riemenschneider14}
Hayko Riemenschneider, Andr{\'a}s B{\'o}dis-Szomor{\'u}, Julien Weissenberg,
  and Luc Van~Gool.
\newblock {Learning Where to Classify in Multi-view Semantic Segmentation}.
\newblock In {\em Proc. ECCV}, 2014.

\bibitem{roynard2018parislille3d}
Xavier Roynard, Jean-Emmanuel Deschaud, and François Goulette.
\newblock {Paris-Lille-3D: A large and high-quality ground truth urban point
  cloud dataset for automatic segmentation and classification}.
\newblock {\em The International Journal of Robotics Research}, 37(6), 2018.

\bibitem{Schult:2020}
Jonas Schult, Francis Engelmann, Theodora Kontogianni, and Bastian Leibe.
\newblock {DualConvMesh-Net: Joint Geodesic and Euclidean Convolutions on 3D
  Meshes}.
\newblock In {\em Proc. CVPR}, 2020.

\bibitem{sepulveda2018nav}
G. Sepulveda, J.~C. Niebles, and A. Soto.
\newblock {A Deep Learning Based Behavioral Approach to Indoor Autonomous
  Navigation}.
\newblock In {\em Proc. ICRA}, 2018.

\bibitem{Song:2017}
Shuran Song, Fisher Yu, Andy Zeng, Angel~X Chang, Manolis Savva, and Thomas
  Funkhouser.
\newblock {Semantic Scene Completion from a Single Depth Image}.
\newblock In {\em Proc. CVPR}, 2017.

\bibitem{Takayama2014Orientation}
Kenshi Takayama, Alec Jacobson, Ladislav Kavan, and Olga Sorkine-Hornung.
\newblock {A} {S}imple {M}ethod for {C}orrecting {F}acet {O}rientations in
  {P}olygon {M}eshes {B}ased on {R}ay {C}asting.
\newblock {\em Journal of Computer Graphics Techniques (JCGT)}, 3(4), 2014.

\bibitem{tan2020toronto3d}
Weikai Tan, Nannan Qin, Lingfei Ma, Ying Li, Jing Du, Guorong Cai, Ke Yang, and
  Jonathan Li.
\newblock {Toronto-3D: A Large-scale Mobile LiDAR Dataset for Semantic
  Segmentation of Urban Roadways}.
\newblock In {\em Proc. CVPR Workshops}, 2020.

\bibitem{Toshev2010}
Alexander Toshev and Ben Taskar.
\newblock Detecting and parsing architecture at city scale from range data.
\newblock In {\em Proc. CVPR}, 2010.

\bibitem{Martinovic2015}
Alexander Toshev and Ben Taskar.
\newblock 3d all the way: Semantic segmentation of urban scenes from start to
  end in 3d.
\newblock In {\em Proc. CVPR}, 2015.

\bibitem{Web:3DWarehouse}
Trimble.
\newblock {\em {3D Warehouse}}, 2020.

\bibitem{uy-scanobjectnn-iccv19}
Mikaela~Angelina Uy, Quang-Hieu Pham, Binh-Son Hua, Duc~Thanh Nguyen, and
  Sai-Kit Yeung.
\newblock Revisiting point cloud classification: A new benchmark dataset and
  classification model on real-world data.
\newblock In {\em Proc. ICCV}, 2019.

\bibitem{Kai19}
Kai Wang, Yu-An Lin, Ben Weissmann, Manolis Savva, Angel~X. Chang, and Daniel
  Ritchie.
\newblock {PlanIT: Planning and Instantiating Indoor Scenes with Relation Graph
  and Spatial Prior Networks}.
\newblock {\em ACM Trans. on Graphics}, 38(4), 2019.

\bibitem{Wang-2020-Unsupervised}
Peng-Shuai Wang, Yu-Qi Yang, Qian-Fang Zou, Zhirong Wu, Yang Liu, and Xin Tong.
\newblock {U}nsupervised 3{D} {L}earning for {S}hape {A}nalysis via
  {M}ultiresolution {I}nstance {D}iscrimination.
\newblock {\em ACM Trans. on Graphics}, 2020.

\bibitem{Wang:2018}
Xiaogang Wang, Bin Zhou, Haiyue Fang, Xiaowu Chen, Qinping Zhao, and Kai Xu.
\newblock {L}earning to {G}roup and {L}abel {F}ine-{G}rained {S}hape
  {C}omponents.
\newblock {\em ACM Trans. on Graphics}, 37(6), 2018.

\bibitem{dgcnn}
Yue Wang, Yongbin Sun, Ziwei Liu, Sanjay~E. Sarma, Michael~M. Bronstein, and
  Justin~M. Solomon.
\newblock Dynamic graph {CNN} for learning on point clouds.
\newblock {\em ACM Trans. on Graphics}, 38(5), 2019.

\bibitem{Web:Wikipedia}
Wikipedia.
\newblock {\em {List of building types}}, 2018.

\bibitem{Li:ShapeNet:2016}
Li Yi, Vladimir~G. Kim, Duygu Ceylan, I-Chao Shen, Mengyan Yan, Hao Su, Cewu
  Lu, Qixing Huang, Alla Sheffer, and Leonidas Guibas.
\newblock {A} {S}calable {A}ctive {F}ramework for {R}egion {A}nnotation in 3{D}
  {S}hape {C}ollections.
\newblock {\em ACM Trans. on Graphics}, 35(6), 2016.

\bibitem{ShapeNetSem}
Li Yi, Lin Shao, Manolis Savva, Haibin Huang, Yang Zhou, Qirui Wang, Benjamin
  Graham, Martin Engelcke, Roman Klokov, Victor~S. Lempitsky, Yuan Gan, Pengyu
  Wang, Kun Liu, Fenggen Yu, Panpan Shui, Bingyang Hu, Yan Zhang, Yangyan Li,
  Rui Bu, Mingchao Sun, Wei Wu, Minki Jeong, Jaehoon Choi, Changick Kim, Angom
  Geetchandra, Narasimha Murthy, Bhargava Ramu, Bharadwaj Manda, M. Ramanathan,
  Gautam Kumar, P. Preetham, Siddharth Srivastava, Swati Bhugra, Brejesh Lall,
  Christian H{\"{a}}ne, Shubham Tulsiani, Jitendra Malik, Jared Lafer, Ramsey
  Jones, Siyuan Li, Jie Lu, Shi Jin, Jingyi Yu, Qixing Huang, Evangelos
  Kalogerakis, Silvio Savarese, Pat Hanrahan, Thomas~A. Funkhouser, Hao Su, and
  Leonidas~J. Guibas.
\newblock {L}arge-{S}cale 3{D} {S}hape {R}econstruction and {S}egmentation from
  {S}hapenet {C}ore55.
\newblock {\em CoRR}, abs/1710.06104, 2017.

\bibitem{Yi2017:syncspec}
Li Yi, Hao Su, Xingwen Guo, and Leonidas~J Guibas.
\newblock {SyncSpecCNN: Synchronized spectral cnn for 3D shape segmentation}.
\newblock In {\em Proc. CVPR}, 2017.

\bibitem{Yu:2019}
Fenggen Yu, Kun Liu, Yan Zhang, Chenyang Zhu, and Kai Xu.
\newblock {P}art{N}et: {A} {R}ecursive {P}art {D}ecomposition {N}etwork for
  {F}ine-{G}rained and {H}ierarchical {S}hape {S}egmentation.
\newblock In {\em Proc. CVPR}, 2019.

\bibitem{zeppelzauer2018age}
Matthias Zeppelzauer, Miroslav Despotovic, Muntaha Sakeena, David Koch, and
  Mario D\"{o}ller.
\newblock Automatic prediction of building age from photographs.
\newblock In {\em Proc. ICMR}, 2018.

\bibitem{Zheng2019}
Jia Zheng, Junfei Zhang, Jing Li, Rui Tang, Shenghua Gao, and Zihan Zhou.
\newblock {Structured3D: A Large Photo-realistic Dataset for Structured 3D
  Modeling}.
\newblock In {\em Proc. ECCV}, 2020.

\bibitem{Zhou:2019}
Yang Zhou, Zachary While, and Evangelos Kalogerakis.
\newblock {SceneGraphNet: Neural Message Passing for 3D Indoor Scene
  Augmentation}.
\newblock In {\em Proc. ICCV}, 2019.

\end{thebibliography}
